\documentclass{article}

\PassOptionsToPackage{numbers, compress}{natbib}

 \usepackage[eandd, final]{neurips_2026}

\usepackage[utf8]{inputenc} 
\usepackage[T1]{fontenc}    
\usepackage{hyperref}       
\usepackage{url}            
\usepackage{booktabs}       
\usepackage{amsfonts}       
\usepackage{nicefrac}       
\usepackage{microtype}      
\usepackage{xcolor}         
\usepackage{pifont}
\usepackage[toc]{multitoc}
\usepackage[skins]{tcolorbox}
\usepackage{graphicx} 
\usepackage{amsmath}
\usepackage{xspace}
\usepackage{wrapfig}

\definecolor{wkred}{RGB}{255, 200, 200}
\definecolor{wkblue}{RGB}{210, 230, 250}
\definecolor{wkgold}{RGB}{255, 223, 129}
\definecolor{wksilver}{RGB}{192, 192, 192}

\definecolor{codegreen}{rgb}{0,0.6,0}
\definecolor{codegray}{rgb}{0.5,0.5,0.5}
\definecolor{codepurple}{rgb}{0.58,0,0.82}
\definecolor{backcolour}{rgb}{0.95,0.95,0.92}
\definecolor{wkgreen}{RGB}{220,244,229}
\definecolor{wkpurple}{RGB}{210,210,253}
\definecolor{wkyellow}{RGB}{255,241,177}

\newcommand{\ccmark}{\textcolor{green}{\ding{51}}}
\newcommand{\cxmark}{\textcolor{red}{\ding{55}}}

\newcommand{\dataft}{\texttt{MultiTalkFT}\xspace}
\newcommand{\datapt}{\texttt{{MultiTalkPT}}\xspace}
\newcommand{\databench}{\texttt{{MultiTalkBench}}\xspace}
\newcommand{\modelname}{\texttt{{Moshi-MTB}}\xspace}

\title{MultiTalk: Scaling Full-Duplex Speech Models to Long, Multi-Party, Bilingual Conversation}

\author{%
  Ke Wang$^{1}$$^{*}$, Houxing Ren$^{1}$$^{*}$, Zimu Lu$^{1}$, Yunqiao Yang$^{1}$,  \\ {\bf  Zhuofan Zong}$^{1}$, {\bf Mingjie Zhan}$^{1}$$^{\dagger}$, {\bf Hongsheng Li}$^{1,2}$$^{\dagger}$\\
  \textsuperscript{1}CUHK MMLab, 
  \textsuperscript{2}CPII under InnoHK \\ 
 \texttt{wangk@link.cuhk.edu.hk} \quad \texttt{hsli@ee.cuhk.edu.hk}
}

\begin{document}

\maketitle

\begin{abstract}
End-to-end full-duplex speech models have brought open-source machine conversation close to human fluency, yet existing systems fall short of real-world deployment in two entangled respects: long-context conversational robustness and multi-party interaction capability. Realistic settings, including meetings, group lessons, family dinners, and social-robot reception, are inherently long-horizon and multi-party at the same time, requiring a single model to perceive, attribute, contextualize, and respond across multiple speakers over extended durations. Progress along these axes is bottlenecked by both data and evaluation. On the data side, open multi-party conversational speech corpora total only a few hundred hours and are not designed for codec-frame-level full-duplex modeling. On the evaluation side, existing long-audio benchmarks focus on passive listening, while existing speech-to-speech benchmarks remain dyadic and short.
In this work, we extend the Moshi paradigm along long-horizon and multi-party axes simultaneously, in both English and Chinese, with three contributions. 
First, we release an open data engine and a \textbf{57.6,k-hour synthetic training corpus} for long, multi-party, English--Chinese full-duplex dialogue. The engine produces parallel-stream audio with controllable length, participant count, conversational dynamics (turn-taking, overlap, backchannels, interruption, addressee shifts, long-range co-reference), and language (English, Chinese), exceeding all prior open multi-party conversational speech corpora by more than an order of magnitude.
Second, we introduce \textbf{MultiTalkBench}, built from real human recordings, to jointly evaluate long-form full-duplex dialogue with an average duration of 32.6 minutes, multi-party, and bilingual full-duplex dialogue, with explicit probes for long-range entity tracking, topic coherence, and addressee selection. 
Third, we train a bilingual Moshi-style full-duplex model on the released corpus that sustains coherent multi-party English--Chinese conversation over extended durations, substantially outperforming open-source baselines including Moshi, MiniCPM-o-4.5, and Qwen3-Omni-30B-A3B-Instruct.
\end{abstract}

\section{Introduction}\label{sec:intro}

End-to-end full-duplex speech models have brought open-source machine conversation closer than ever to human fluency. A single model now listens, thinks, and speaks simultaneously, collapsing the long-standing ASR\,$\rightarrow$\,LLM\,$\rightarrow$\,TTS pipeline into a unified system with sub-second response latency, in line with commercial reference points such as GPT-4o~\citep{openai2024gpt4ocard} and Gemini Live~\citep{geminiteam2025geminifamilyhighlycapable,comanici2025gemini25pushingfrontier}. Among open systems, Moshi~\citep{kyutai2024moshi} has emerged as the de facto architectural reference. Its ideas have been adopted, refined, or competed against by a rapidly growing family of full-duplex and streaming speech-to-speech models, including J-Moshi~\cite{ohashi2025jmoshi}, GLM-4-Voice~\cite{zeng2024glm4,zeng2024scaling}, Step-Audio~\cite{huang2025stepaudio,wu2025stepaudio2}, Qwen-Omni~\cite{xu2025qwen25omni,xu2025qwen3omni}, Kimi-Audio~\cite{kimiteam2025kimiaudio}, MiniCPM-o~\citep{yao2024minicpm}, and bilingual native full-duplex systems such as FLM-Audio and RoboEgo~\cite{yao2025flmaudio,yao2025roboego}.

Despite this rapid progress, existing full-duplex models fall short of real-world deployment in two critical respects: long-context conversational robustness and multi-party interaction capability. A deployed conversational agent runs continuously rather than over short isolated clips, so realistic settings, including meetings, group lessons, family dinners, and social-robot reception scenarios, are inherently long-horizon and multi-party at the same time. Across hours of continuous audio, a single model must simultaneously perceive utterances from multiple distinct human speakers, attribute each utterance to the correct speaker, maintain coherent context over extended history, and respond with appropriate addressee selection. This is fundamentally a problem of semantics and turn-taking, orthogonal to the acoustic source-separation and diarization literature. Classical multi-party dialogue research~\cite{skantze2021turntaking,ekstedt2022vap,inoue2024vap}, addressee detection~\cite{zhang2018addressee,gu2021mpcbert,inoue2025addressee}, and recent multi-party social-robot work~\cite{abbo2025multiparty} have advanced modular pipelines. However, no end-to-end full-duplex speech model has been trained or evaluated for the one-model, many-user, long-horizon setting that real deployment demands.

\begin{wrapfigure}{r}{0.5\textwidth}
    \centering
    \vspace{-\baselineskip}
    \includegraphics[width=0.48\textwidth]{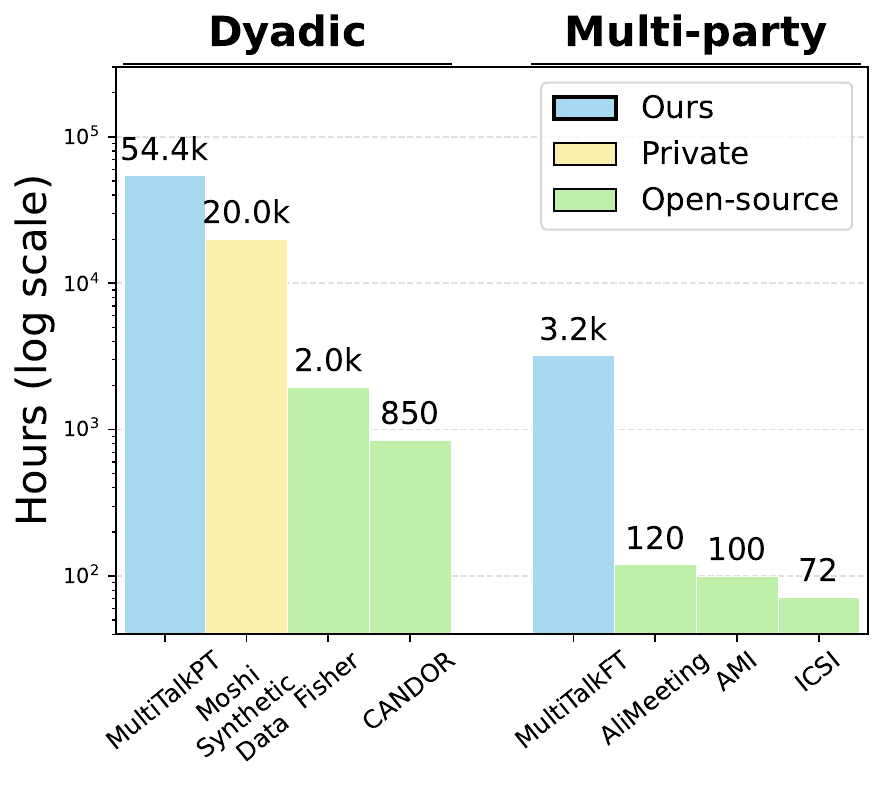}
    \vspace{-5mm}
    \caption{Training-data scale comparison.}
    \label{fig:data_comparison}
\end{wrapfigure}

Realizing this capability depends on the availability of suitable training data, yet such data is extremely difficult and costly to collect. An open-source spoken-dialogue corpus for long\,$\times$\,multi-party\,$\times$\,bilingual\,$\times$\,parallel-stream training is essentially absent. Classical dyadic telephone corpora, including Switchboard~\cite{godfrey1992switchboard}, Fisher~\cite{cieri2004fisher}, and HKUST/MTS~\cite{liu2006hkust}, are monolingual. Open multi-party speech corpora, including AMI~\cite{carletta2006ami}, ICSI~\cite{janin2003icsi}, AISHELL-4~\cite{fu2021aishell4}, and AliMeeting~\cite{yu2022alimeeting}, together amount to only a few hundred hours and were not designed for codec-frame-level full-duplex modeling. Reflecting this scarcity, every recent Moshi-style system falls back on proprietary in-house data or on TTS-synthesized stereo dialogue that is not publicly redistributed~\cite{kyutai2024moshi,veluri2024syncllms,zhang2025omniflatten,zeng2024glm4,huang2025stepaudio,ohashi2025jmoshi}. An open data engine for long, multi-party, bilingual full-duplex audio is precisely the missing ingredient.

Compounding these training-data gaps, current evaluation protocols are also inadequate as shown in Table~\ref{tab:multitalk-stats}. Recent long-audio benchmarks, including BLAB~\cite{ahia2025blabbrutallylongaudio}, AudioMarathon~\cite{he2025audiomarathoncomprehensivebenchmarklongcontext}, and ChronosAudio~\cite{luo2026chronosaudio}, demonstrate that audio LLMs suffer substantial accuracy degradation as audio length grows from seconds to tens of minutes. Crucially, however, these benchmarks measure \textit{passive} listening rather than interactive generation. Speech-to-speech benchmarks such as VoiceBench, AudioBench, AIR-Bench, SD-Eval, and S2S-Arena focus on dyadic instruction-following, whereas Talking Turns, Full-Duplex-Bench, and MTalk-Bench focus on dyadic turn-taking and overlap dynamics. URO-Bench~\cite{yan2025urobench} is the only speech-to-speech benchmark that covers bilingual English--Chinese multi-round interaction alongside paralinguistic dimensions, yet it too remains dyadic and short.

Meaningful progress along any of these dimensions requires a unified treatment that integrates model, training corpus, and benchmark. In this work, we extend the Moshi paradigm along both axes simultaneously and in both Chinese and English. Our contributions are as follows:

\begin{itemize}
    \item \textbf{A data engine and an open 57.6,k-hour synthetic training corpus for long, multi-party, bilingual full-duplex dialogue.} The engine generates parallel-stream audio conversations with controllable length, number of participants, conversational dynamics (turn-taking, overlap, backchannels, interruption, addressee shifts, long-range co-reference), and language (English, Chinese). It combines persona- and memory-conditioned dialogue planning, multi-speaker bilingual TTS, and acoustically faithful multi-channel mixing to produce data with codec-frame-level alignment suitable for Moshi-style training. We release both the engine and a 57.6\,k-hour corpus, exceeding the union of all prior open full-duplex speech corpora as shown in Figure~\ref{fig:data_comparison}.

    \item \textbf{MultiTalkBench, the first benchmark to jointly evaluate long, multi-party, and bilingual full-duplex dialogue.} MultiTalkBench tests speech-to-speech systems on (a) interactive conversations longer than ten minutes with explicit probes for long-range entity tracking and topic coherence, (b) one-model-many-user multi-party interaction with quantitative addressee-selection and turn-taking metrics, and (c) English--Chinese bilingual abilities. To our knowledge, no prior benchmark addresses these axes jointly in a fully interactive,
    end-to-end speech setting.

    \item \textbf{A bilingual Moshi-style full-duplex model.} The model demonstrates that a single end-to-end system can sustain coherent multi-party English--Chinese conversation over extended durations, substantially outperforming open-source baselines including Moshi, MiniCPM-o-4.5, and Qwen3-Omni-30B-A3B-Instruct.
\end{itemize}

\begin{table*}[t]
  \centering
  \caption{Statistics of \databench\ and prior speech evaluation benchmarks.}
  \resizebox{0.7\textwidth}{!}{%
  \label{tab:multitalk-stats}
  \begin{tabular}{l|c|ccc}
    \toprule
     
     & \textbf{\databench}
     & \textbf{VoiceBench} & \textbf{FDB-v1.5} & \textbf{MTalk-Bench} \\
    \midrule
Total (h)         & 56.5    & 55.9     & 1.9      & 1.5 \\
Max. (min)        & 43.6    & 0.9      & 0.2      & 0.8 \\
    \midrule
    Multi-party           & \ccmark & \cxmark  & \cxmark  & \ccmark\; (only 9.8 min) \\
    Full-duplex             & \ccmark & \cxmark  & \ccmark  & \cxmark \\
    Bilingual          & \ccmark & \cxmark  & \cxmark  & \cxmark \\
    \bottomrule
  \end{tabular}
  }
\end{table*}

\section{Methods}\label{sec:methods}
\subsection{Automatic Data Engine}\label{sec:data_engine}
\begin{figure*}[t]
    \centering
    \includegraphics[width=1.67\textwidth]{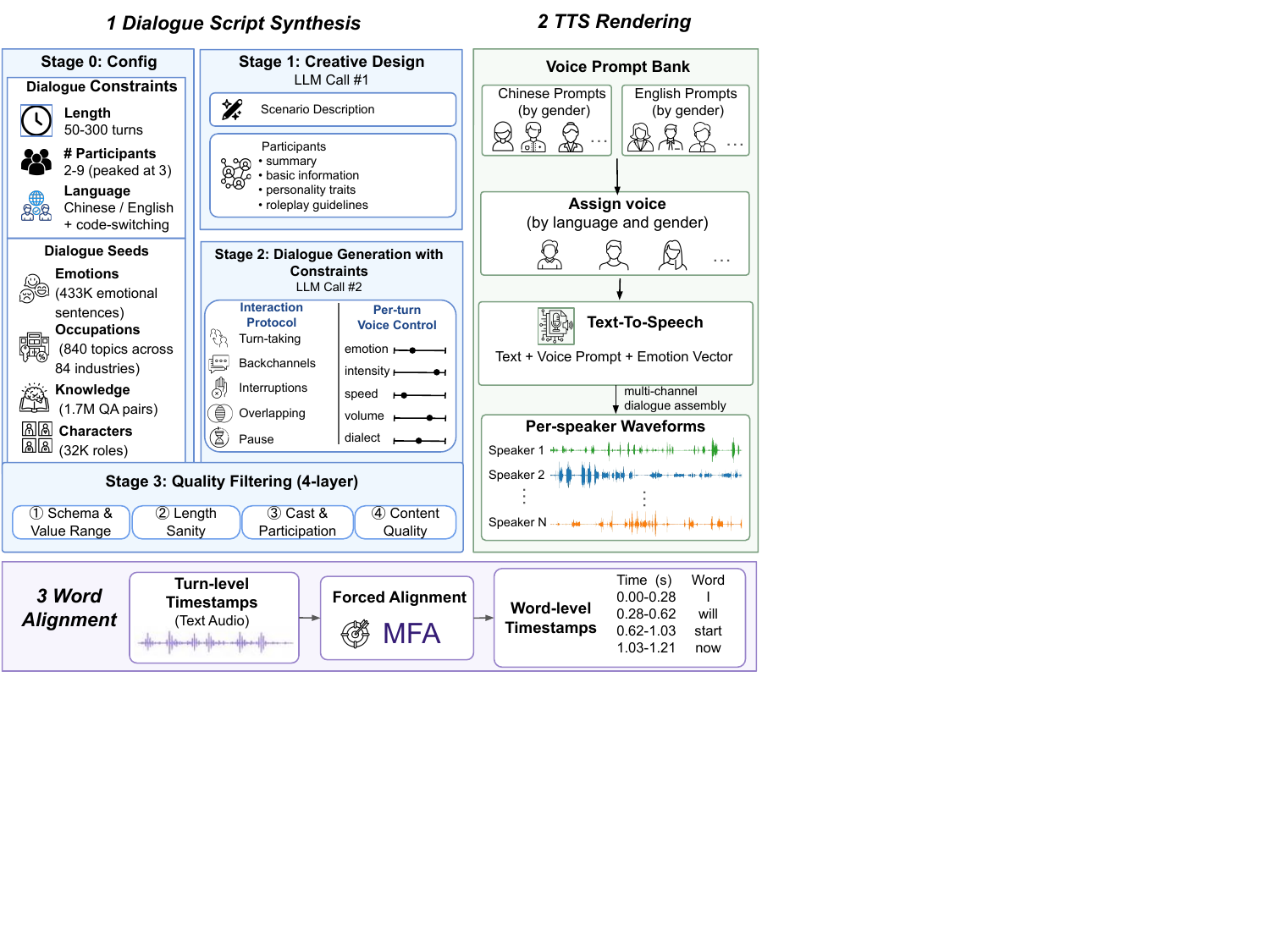}
    \vspace{-56mm}
    \caption{\textbf{Automatic data engine for parallel-stream full-duplex dialogue.} \textbf{(1) Dialogue Script Synthesis} turns a sampled configuration (length, participants, language) and four seed pools (emotions, occupations, knowledge, characters) into a constrained dialogue script via two LLM calls followed by a four-layer quality filter. \textbf{(2) TTS Rendering} synthesizes per-utterance waveforms with IndexTTS2 from a gender-matched voice prompt and an emotion vector, then assembles them into per-speaker parallel streams. \textbf{(3) Word Alignment} refines turn-level timestamps to word-level alignments via Montreal Forced Aligner (MFA).}
    \label{fig:data_engine}
    \vspace{-2mm}
\end{figure*}

To construct long, multi-party, bilingual full-duplex dialogue data for Moshi-style training, we build an automatic data engine (Figure~\ref{fig:data_engine}) that produces parallel-stream speech along three controllable axes: conversation length, number of participants, and language. Given dialogue seeds spanning emotions, occupations, knowledge, and characters, the engine first performs two-pass symbolic script synthesis: one LLM call constructs the scenario, cast, and interaction trajectory, while a second call realizes the full dialogue under this fixed world model. The scripts are then rendered with bilingual multi-speaker TTS and multi-channel mixing, followed by word-level alignment for codec-frame-level supervision.


\subsubsection{Seed Data Collection}\label{sec:seed}

To support diverse and grounded dialogue generation, we assemble four pools of seed material that serve as optional conditioning input for the script synthesizer, including emotional seed, knowledge seed, character seed, and occupational seed. 
Emotional seeds are taken from \texttt{dair-ai/emotion}~\cite{saravia2018carer}, yielding 433K sentences annotated with discrete affect labels for grounding emotional tone. 
Knowledge seeds consist of 1.7M question--answer pairs from \texttt{AM-DeepSeek-R1-0528-Distilled} and \texttt{AM-Qwen3-Distilled}~\cite{tian2025amdistillation}, which together span a broad spectrum of reasoning domains; we apply keyword-based filtering to discard items containing code fragments, markdown artifacts, error tracebacks, and other formatting unsuitable for spoken interaction. 
Character seeds are obtained by prompting \texttt{gemini-2.5-pro} to normalize raw role descriptions into a unified schema covering identity, background, personality traits, and roleplay guidelines, producing 32K deduplicated profiles. 
Occupational seeds comprise 840 discussion topics organized across 84 industries, supplying domain-specific context for task-oriented conversation.
Synthetic-data construction and back-translation have also been explored in mathematical reasoning and software-development tasks~\cite{lu2024mathgenie,lu2025mathcoder2,lu2026fullstackagent,lu2026webgenagent}.

\subsubsection{Dialogue Script Synthesis}\label{sec:script}

Each dialogue is conditioned on a small set of seeds together with a sampled scenario configuration that fixes its global parameters: number of participants, ranging from 2 to 9 and most commonly 3, target turn count, ranging from 50 to 300, language (English or Chinese), and level of AI involvement. Seeds are injected via prompt-level system-message substitution. To separate global scenario planning from local dialogue realization, we use a two-stage synthesis pipeline. The first call constructs the world model, including the cast, setting, participant roles, and high-level interaction trajectory. The second call realizes the full dialogue under this fixed scenario, while enforcing per-turn TTS annotations, speaker consistency, and cast closure.

\paragraph{Creative Design.}
The first call emits a creative-design block, a four- to six-sentence scenario paragraph, and a closed cast in which each participant is described along four \emph{stable} dimensions: summary, basic information, personality traits, and roleplay guidelines. Transient affective state is confined to the scenario paragraph, so that the second call can modulate per-turn TTS controls without contradicting any persona. The prompt enforces a \emph{cast-closure} invariant (every named participant in the scenario must appear in the participant list) and instructs the model not to quote any seed verbatim, breaking the surface-form attractor we observed in single-prompt formulations.

\paragraph{Dialogue Generation.}
The second call ingests this creative-design block and emits an array of turns subject to three constraint families encoded directly in the prompt. A \emph{spoken-length distribution} requests 60\% short turns (1--15 words), 30\% medium (15--40), and 10\% long, with consecutive long turns prohibited. An \emph{interaction protocol} schematizes four behaviors: backchannels are short turns inserted between another speaker's adjacent turns; interruptions terminate the preceding turn with an em-dash marker, followed immediately by the interrupting turn; overlapping speech is flagged on adjacent turns; and pauses are realized as explicit silence turns $\{\text{speaker}{=}\text{silence},\ \text{text}{=}[Xs]\}$ with $X\in[1,5]$. Explicit silence turns are critical for full-duplex training because they let the renderer insert genuine acoustic silence rather than rely on inter-turn gaps that vanish under tight TTS concatenation. Finally, a \emph{five-dimensional TTS control vector} (emotion, intensity, speed, volume, dialect) is attached to each non-silence turn, and its \texttt{text} field is constrained to be \emph{TTS-safe}: parenthetical or bracketed performance cues are forbidden, all affective information is carried by the voice control object, and the field is passed verbatim to the TTS module.
More broadly, code-assisted reasoning and reflection provide examples of structured intermediate generation and verification in language models~\cite{zhou2024gpt4codeinterpreter,wang2024mathcoder,ren2025reflectioncoder}.

\paragraph{Quality Filtering.}
A four-layer filter rejects scripts with structured reason codes, so that prompt iterations can be tied to per-class failure-rate deltas. The filter verifies (i) JSON schema and TTS-annotation value ranges, (ii) spoken-length sanity (per-turn word count and consecutive same-speaker runs), (iii) cast closure and minimum participation per speaker, and (iv) content quality (no adjacent verbatim repetition, no AI-template openings, no role-breaking phrases).
Other generation methods use editing, infilling, or alignment objectives to improve outputs; these are complementary methodological directions rather than components of our data engine~\cite{ren2026editbasedrefinement,ren2024characterleveltextinfilling,ren2025fillinthemiddle}.

\subsubsection{TTS Rendering and Word Alignment}\label{sec:tts}

\paragraph{Per-Utterance Synthesis.}
Each script is rendered into per-utterance waveforms by IndexTTS2~\cite{zhou2025indextts2}, a zero-shot TTS model conditioned on a speaker prompt and an emotion vector. We curate Chinese and English prompt banks partitioned by gender. For each dialogue, the bank matching the script language is shuffled, and prompts are drawn without replacement and assigned to participants by gender, ensuring that no two participants share a voice. Drawing prompts at the dialogue level rather than the corpus level yields exponentially many prompt combinations per cast, breaking the speaker-text correlations that would otherwise let downstream models exploit shortcuts for speaker identification. For each non-silence turn, the synthesizer receives an eight-dimensional emotion vector obtained by one-hot indexing the annotated emotion category and scaling by its intensity. Silence turns bypass the synthesizer and emit zero-valued audio of the requested duration.

\paragraph{Multi-Channel Assembly.}
The per-utterance waveforms are assembled into a multi-channel track in which the assistant occupies channel 0, producing the per-speaker waveform pair consumed by downstream models. Each turn is mixed into its speaker's channel at a position determined by sequential placement, subject to a same-channel non-overlap invariant that forbids any turn from beginning before the previous endpoint of its own channel. Cross-speaker overlap, by contrast, is freely admitted, and the script's interaction protocol is realized through this asymmetric placement policy. A subsequent gap-compression pass tightens inter-turn silence so that adjacent cross-speaker boundaries overlap by 0.2 to 0.6 seconds for normal transitions, or by 1 to 2 seconds when the preceding turn was marked as interrupted.

\paragraph{Word Alignment.}
To support training a time-aligned text stream in the spirit of Moshi's Inner Monologue, we obtain word-level timestamps for assistant turns by forced alignment with the Montreal Forced Aligner~\citep{mcauliffe2017mfa}.

\subsection{MultiTalkBench}\label{sec:multitalkbench}

\begin{figure*}[t]
    \centering
    \includegraphics[width=1.07\textwidth]{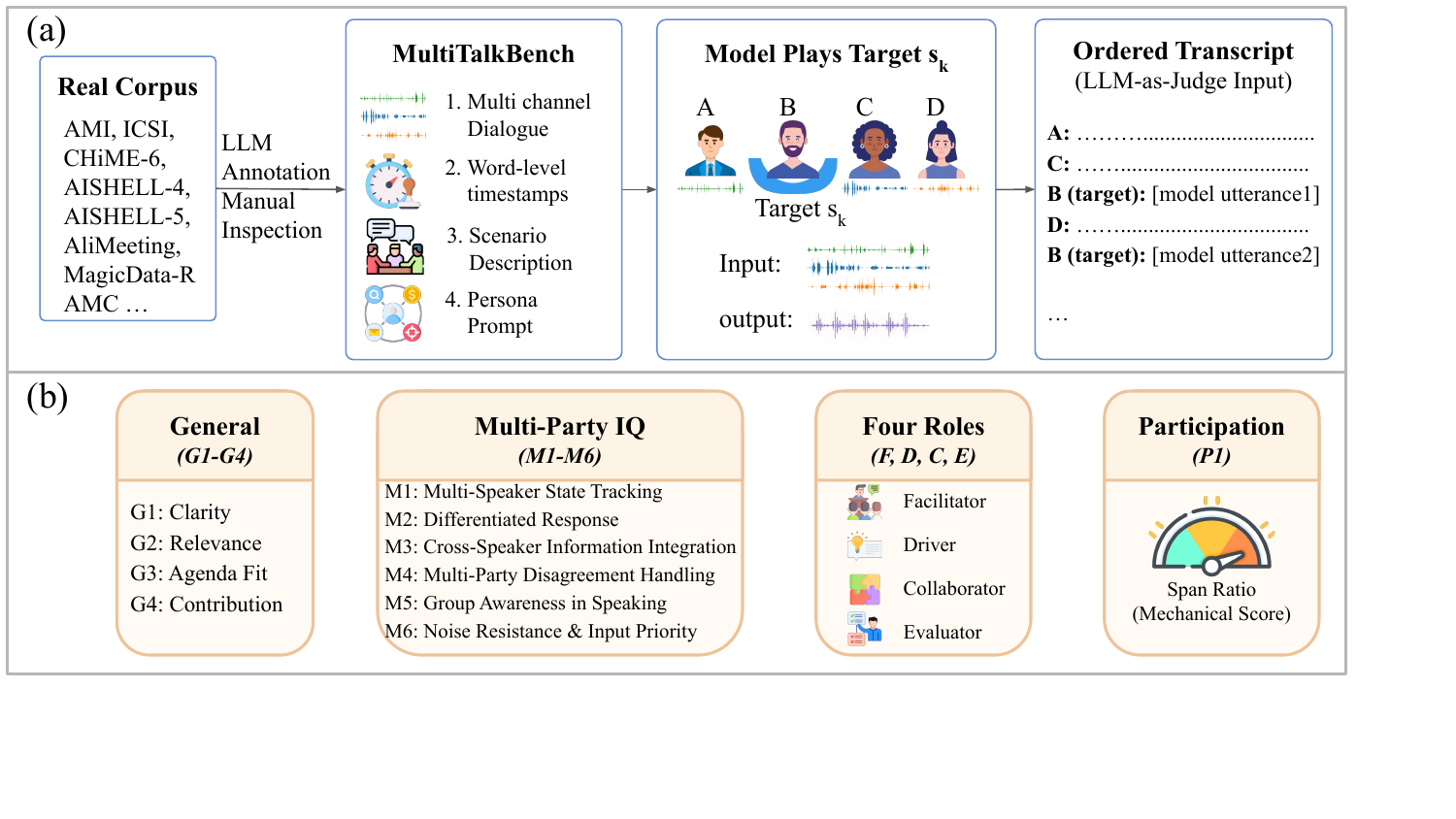}
    \vspace{-20mm}
\caption{\textbf{MultiTalkBench.} \textbf{(a)} Each test instance is built from a real multi-party conversation annotated and inspected into multi-channel dialogue with word-level timestamps, a scenario description, and per-speaker persona prompts. The model plays a designated target participant $s_k$, and its generated utterances are spliced into the ordered transcript fed to an LLM-as-Judge. \textbf{(b)} The metric spans four general dimensions, six multi-party intelligence dimensions, four roles, and a mechanical span-ratio participation score.}
\label{fig:multitalkbench}
    \vspace{-2mm}
\end{figure*}

To evaluate full-duplex speech models under realistic long-form interaction, we introduce \textbf{MultiTalkBench} (Figure~\ref{fig:multitalkbench}), a benchmark for long, multi-party, bilingual dialogue. MultiTalkBench places the model in a multi-party meeting as a \emph{specific named participant}, rather than as a generic one-on-one assistant. This design probes capabilities that are central to deployed conversational agents but under-covered by existing benchmarks: long-range entity tracking, topic coherence, addressee selection, and turn-taking across multiple speakers.


\paragraph{Data Collection.}
Unlike the synthetic training corpus, MultiTalkBench is constructed from real human conversational recordings.
The audio is sourced from the test split of publicly released multi-party speech corpora that together span English meeting (AMI~\cite{carletta2006ami}, ICSI~\cite{janin2003icsi}), English naturalistic dinner-party (CHiME-6~\cite{watanabe2020chime6}), Mandarin meeting (AISHELL-4~\cite{fu2021aishell4}, AliMeeting~\cite{yu2022alimeeting}), Mandarin in-car multi-speaker (AISHELL-5~\cite{aishell5}), and Mandarin conversational telephone (MagicData-RAMC~\cite{yang2022magicdata}) settings. We retain only sessions for which per-speaker close-mic or lapel audio is available, since far-field arrays alone do not yield the channel-isolated streams required to feed each non-target participant into the model as a separate input. Sessions are further filtered for clean alignment, absence of ghost speakers, and per-channel SNR above an acoustic-quality threshold. From the filtered pool we generate 104 evaluation samples. 
We construct a persona prompt per $(M, s_k)$ via a two-pass offline pipeline. Per-speaker seeds (alias, self-introduced name when available, gender heuristic, role hint, three representative quotes) are sent with a transcript excerpt to a writer LLM, which returns a meeting-level scenario paragraph plus a 10-field profile per speaker.
Source corpora such as AMI provide audio but define no model task. MultiTalkBench reuses them as input audio while contributing the task definition, persona prompts, dimensions, and scoring procedure.

\paragraph{Metric.}

The metric covers four groups. The \textit{General} group evaluates basic response quality, including clarity, relevance, agenda fit, and contribution. The \textit{Multi-Party IQ} group evaluates capabilities specific to conversations with $\geq 3$ speakers, including multi-speaker state tracking, differentiated response to different addressees, cross-speaker information integration, multi-party disagreement handling, group awareness in speaking, and resistance to noisy or low-priority inputs. The \textit{Role-conditional} group evaluates role-specific behavior for four meeting roles: Facilitator, Driver, Collaborator, and Evaluator. Finally, \textit{Participation} is a mechanical dimension defined below.
For target speaker $s_k$ in meeting $M$ with utterance set $U_k \subseteq U$, we define the time-\emph{span} participation ratio and its matching coefficient as
\begin{equation}
r(s_k) = \frac{\max_{u \in U_k} u.\text{end} - \min_{u \in U_k} u.\text{start}}{\max_{u \in U} u.\text{end} - \min_{u \in U} u.\text{start}},
\quad
\mathrm{coef}(s_k) = \max\bigl(0,\, 1 - |r_{\text{model}} - r^{\text{GT}}|\bigr) \in [0,1],
\end{equation}
where $r^{\text{GT}}$ is computed identically over the human reference alignment, and we report $P_1 = 100\cdot\mathrm{coef} \in [0,100]$.
For each sample, the target $s_k$ is assigned exactly one of the four meeting roles, so within the Role-conditional group only that role's sub-dimensions are scored. The four roles therefore partition the sample set, and the role-group total is the \emph{sum} of the four per-role scores, in contrast to the General and Multi-Party IQ groups, whose sub-dimensions apply to every sample and \emph{average} to the group total. The three LLM-judge groups are weighted equally ($1/3$ each) in the final score.
These content-level scores are assigned to offline model responses inserted into the human-conversation transcript; they do not directly measure response latency or interruption timing.

\subsection{Training}\label{sec:training}
To obtain a Moshi-style model capable of long-horizon, multi-party, bilingual full-duplex dialogue, we adopt a two-phase training recipe initialized from the public \texttt{kyutai/moshiko-pytorch-bf16}. In the first phase, we perform bilingual pre-training on \datapt, a 54.4k-hour full-duplex corpus comprising both English and Chinese dialogues (Table~\ref{tab:multitalk-stats}). Per-language sampling weights are tuned so that each epoch is approximately 60\% English and 40\% Chinese, adapting the model to Chinese while preserving English ability and introducing a user-stream prediction objective. In the second phase, we specialize the resulting dyadic checkpoint for multi-party interaction by fine-tuning on \dataft, a 3.2k-hour bilingual multi-party corpus with per-speaker streams and system prompts. To isolate the contribution of data, we leave Moshi's dual-stream architecture entirely unmodified and mix all non-target speakers into the user channel. This stage restricts training to multi-party conversations so that the model concentrates capacity on cross-talk with multiple users.




\section{Experiments}\label{sec:experiments}

In this section, we evaluate the effectiveness of our data engine (Section~\ref{sec:data_engine}) and two-phase training recipe (Section~\ref{sec:training}) for long, multi-party, bilingual full-duplex dialogue. We compare \modelname with four publicly available speech dialogue baselines on \databench (Section~\ref{sec:multitalkbench}), focusing on both long-context conversational consistency and multi-party interaction ability.


\subsection{Implementation Details}\label{implementation_details}

\paragraph{Parameters.}
Both training phases share the same loss weighting and user-stream design. The text head uses a padding weight of $0.2$ and an end-of-text padding weight of $0.6$. The audio heads use a first-codebook weight multiplier of $100$ and a non-semantic-codebook weight of $1.0$ on the Moshi stream. The user audio stream is supervised jointly with the Moshi stream: its loss is scaled by $0.5$, with a first-codebook multiplier of $100$ and a non-semantic-codebook weight of $0.5$, and is linearly warmed up over the first $500$ optimizer steps to avoid destabilizing the pre-trained Moshi-side distribution. On-the-fly augmentation mixes far-field noise sampled from the DNS-Challenge noise corpus into the user channel.
We perform full-parameter training with AdamW and a one-cycle schedule with $\text{pct\_start}{=}0.01$. Both phases use a global batch size of $21$\,h of audio.
For pre-training, the model is trained on \datapt for $5{,}000$ steps. The peak learning rate is $3\!\times\!10^{-5}$.
For fine-tuning, the model is trained exclusively on \dataft for $200$ steps, with the peak learning rate lowered to $2\!\times\!10^{-6}$ for the temporal transformer and $4\!\times\!10^{-6}$ for the depth transformer.

\paragraph{Baselines.}
\textbf{Moshiko-7B}~\cite{kyutai2024moshi} is the original Moshi base without persona finetuning. \textbf{PersonaPlex-7B}~\cite{nvidia2026personaplex} is a persona-conditioned Moshi variant. \textbf{MiniCPM-o-4.5}~\cite{yao2024minicpm} is a streaming chunked model with explicit \texttt{is\_listen} gating. \textbf{Qwen3-Omni-30B}~\cite{xu2025qwen3omni} is a 30B mixture-of-experts foundation model, the largest in the comparison. \modelname (ours) is a Moshi-7B backbone finetuned on the multi-party portion of our corpus with the language-balanced training recipe in Section~\ref{sec:training}.

\begin{table*}[t]
\centering
\caption{Detailed results on \databench, broken down by sub-dimension. \textbf{Bold}: best result among models per row. The $\Delta$ column reports relative change of Moshi-MTB over its Moshiko-7B backbone. \textit{Italics}: human reference (oracle upper bound).}
\label{tab:main_results}
\setlength{\tabcolsep}{4pt}
\resizebox{\textwidth}{!}{%
\begin{tabular}{l|cc|cc|c|c|c}
\toprule
& \multicolumn{2}{c|}{\textit{Open SOTA}} & \multicolumn{2}{c|}{\textit{Moshi family}} & & & \\
 & Qwen3-Omni & MiniCPM-o-4.5 & PersonaPlex & Moshiko-7B & \textbf{Moshi-MTB} & $\Delta$ & \textit{Human} \\
\midrule
Size & 30B & 9B & 7B & 7B & 7B & -- & -- \\
\midrule
\textbf{General}                       & 6.93           & 16.35          & 2.76           & 4.96          & \textbf{18.73}          & $+277.6\%$    & \textit{80.53} \\
\quad Clarity & 5.91           & 18.63          & 4.35           & 5.25          & \textbf{19.41}          & $+269.7\%$    & \textit{69.71} \\
\quad Relevance & 6.47           & 21.91          & 3.23           & 7.52          & \textbf{22.72}          & $+202.1\%$    & \textit{88.22} \\
\quad Agenda Fit & 8.52           & 14.88          & 3.46           & 6.58          & \textbf{25.39}          & $+285.9\%$    & \textit{89.42} \\
\quad Contribution & 6.80           & \textbf{9.97}  & 0.00           & 0.48          & 7.40                    & $+1441.7\%$   & \textit{74.76} \\
\midrule
\textbf{Multi-Party IQ}                & 2.81           & \textbf{9.00}  & 1.71           & 2.41          & 8.56                    & $+255.2\%$    & \textit{64.18} \\
\quad Speaker Tracking & 3.89           & 13.60          & 4.45           & 5.41          & \textbf{14.56}          & $+169.1\%$    & \textit{82.45} \\
\quad Differentiated Response & 2.28           & \textbf{5.85}  & 1.20           & 1.66          & 4.37                    & $+163.3\%$    & \textit{54.57} \\
\quad  Information Integration & 3.78           & \textbf{9.38}  & 1.84           & 1.44          & 2.89                    & $+100.7\%$    & \textit{63.94} \\
\quad Disagreement Handling & 0.90           & \textbf{2.54}  & 0.00           & 0.00          & 0.96                    & --  & \textit{40.62} \\
\quad Group Awareness & 5.23           & 14.40          & 2.06           & 3.58          & \textbf{15.80}          & $+341.3\%$    & \textit{70.43} \\
\quad Noise Resistance & 0.78           & 8.23           & 0.72           & 2.39          & \textbf{12.77}          & $+434.3\%$    & \textit{73.08} \\
\midrule
\textbf{Role-conditional}                          & 5.08           & 7.81           & 3.10           & 6.62          & \textbf{12.16}          & $+83.7\%$     & \textit{61.51} \\
\quad Facilitator                      & 1.24           & 0.51           & 0.67           & 0.71          & \textbf{5.11}           & $+619.7\%$    & \textit{16.20} \\
\quad Driver                           & \textbf{3.54}  & 1.90           & 0.66           & 0.30          & 1.49                    & $+396.7\%$    & \textit{33.35} \\
\quad Collaborator                     & 0.30           & 4.84           & 1.70           & \textbf{5.55} & 5.24                    & $-5.6\%$      & \textit{9.50}  \\
\quad Evaluator                        & 0.00           & \textbf{0.56}  & 0.06           & 0.06          & 0.33                    & $+450.0\%$    & \textit{2.46}  \\
\midrule
\textbf{Participation}                 & 13.71          & 37.10          & \textbf{97.87} & 97.86         & 91.94                   & $-6.0\%$      & \textit{100.00} \\
\midrule
\textbf{Final Score}                         & 4.94           & 11.05          & 2.52           & 4.66          & \textbf{13.15}          & $+182.2\%$    & \textit{68.74} \\
\quad EN                          & 4.54           & 7.25           & 4.27           & 7.82          & \textbf{10.87}          & $+39.0\%$     & \textit{68.35} \\
\quad ZH                          & 5.49           & 16.23          & 0.14           & 0.36          & \textbf{16.27}          & $+4419.4\%$   & \textit{69.28} \\
\bottomrule
\end{tabular}%
}
\end{table*}

\subsection{Main Results}\label{sec:main_results}

Table~\ref{tab:main_results} reports whether the data engine and corpus introduced in Section~\ref{sec:data_engine} provide effective training material for the long, multi-party, bilingual full-duplex setting. First, training the off-the-shelf Moshiko-7B checkpoint on $\datapt + \dataft$ under our two-phase recipe increases the overall final score by $+182.2\%$ (Moshiko-7B $4.66 \to$ \modelname $13.15$). Second, the resulting model outperforms all four open-source baselines, including MiniCPM-o-4.5 ($11.05$) and Qwen3-Omni-30B ($4.94$). Together, these observations provide consistent evidence for the effectiveness of the proposed training data.

\paragraph{Multi-party gain.}
The two groups that target multi-party competence both improve sharply over the Moshiko-7B backbone. Multi-Party IQ rises from $2.41$ to $8.56$ ($+255.2\%$), narrowly trailing the strongest baseline (MiniCPM-o-4.5, $9.00$). \modelname in fact leads MiniCPM-o-4.5 on Speaker Tracking ($14.56$ vs.\ $13.60$), Group Awareness ($15.80$ vs.\ $14.40$), and Noise Resistance ($12.77$ vs.\ $8.23$), with the residual aggregate gap concentrated on Information Integration and Disagreement Handling. The Role-conditional group, which scores behaviors that emerge only when the model is assigned a designated meeting role, increases from $6.62$ to $12.16$ ($+83.7\%$), exceeding MiniCPM-o-4.5 ($7.81$) by $+55.7\%$. The largest model in the comparison, Qwen3-Omni-30B, reaches only $2.81$ and $5.08$ on these two groups respectively, indicating that performance on multi-party behavior is bounded by training distribution rather than by parameter count. This empirically validates the multi-party diagnosis of Section~\ref{sec:intro}: existing open full-duplex models are trained predominantly on dyadic interaction, and one operative intervention is data composition.

\paragraph{Bilingual unlock.}
The largest single improvement occurs on Chinese. ZH final score increases from $0.36$ on Moshiko-7B to $16.27$ on \modelname, a factor of about $45\times$, converting an English-only checkpoint into a bilingual model. English performance is preserved, with EN final increasing from $7.82$ to $10.87$ (+39.0\%). This pattern is consistent with the bilingual-by-design construction of \datapt, which introduces a second language without displacing the first.

\paragraph{Limitations and headroom.}
On Multi-Party IQ, MiniCPM-o-4.5 narrowly leads ($9.00$ vs.\ $8.56$), which we attribute to its broader pre-training corpus. On Participation, PersonaPlex-7B attains the highest score ($97.87$), yet its General, Multi-Party IQ, and Role-conditional scores all fall in the bottom quartile, yielding a final score of $2.52$, the lowest among the five compared models. All five models remain far below the human reference ($13.15$ vs.\ $68.74$), indicating substantial headroom on the long, multi-party, bilingual full-duplex setting evaluated here.

\paragraph{Additional analyses.}
With the preceding training stage fixed, increasing the fraction of
\dataft from 0\% to 25\%, 50\%, 75\%, and 100\% raises the Final Score
from 7.95 to 10.39, 11.78, 12.77, and 13.15, respectively. In
size-matched 800-hour comparisons, the score is 10.39 with a random
subset, 8.30 with only three-speaker conversations, and 9.55 with only
the shortest conversations. \modelname scores 13.44 on three-speaker
samples versus 11.96 on larger groups, and 21.23 on conversations under
30 minutes versus 12.07 on longer ones. The benefit also transfers to
F-Actor~\citep{züfle2026factor}, whose Final Score increases from 6.94 to 15.16 after training
with MultiTalk. To complement \databench's offline content scores,
we directly measure interaction behavior: \modelname achieves 870\,ms
response onset latency, 2230\,ms stop latency, 94.1\% backchannel
continuation, and 98.8\% side-conversation ignoring, versus 782\,ms,
5475\,ms, 93.7\%, and 92.0\% for Moshiko-7B. These results support
benefits from \dataft scale and diversity, cross-architecture
transfer, and improved interruption handling, without establishing a
scaling law for the entire corpus or isolating every source of
synthetic-to-real variation.

\subsection{Human Evaluation}\label{sec:human_evaluation}

\begin{table}[h]
    \centering
    \caption{Correlation between human references and two LLM judges (Gemma-4-31B and Qwen3.5-27B) on \databench. Each cell reports Spearman's $\rho$ / Kendall's $\tau$.}
    \label{tab:human-eval}
\resizebox{\textwidth}{!}{%
    \begin{tabular}{l|c|c|c}
        \toprule
        \textbf{Metric Group} & \textbf{Gemma-4-31B vs.\ Human} & \textbf{Qwen3.5-27B vs.\ Human} & \textbf{Gemma-4-31B vs.\ Qwen3.5-27B} \\
                              & ($\rho$ / $\tau$)         & ($\rho$ / $\tau$)        & ($\rho$ / $\tau$) \\
        \midrule
        General             & 0.845 / 0.739 & 0.761 / 0.614 & 0.838 / 0.701 \\
        Multi-Party IQ      & 0.854 / 0.764 & 0.770 / 0.644 & 0.768 / 0.649 \\
        Role-conditional    & 0.888 / 0.805 & 0.734 / 0.631 & 0.720 / 0.604 \\
        \midrule
        Final Score         & 0.898 / 0.787 & 0.846 / 0.700 & 0.847 / 0.706 \\
        \bottomrule
    \end{tabular}
    }
\end{table}

To validate that the LLM-as-judge protocol of Section~\ref{sec:multitalkbench} produces scores aligned with human judgment, we conduct a human evaluation comparing rater scores against two LLM judges. We randomly sample 20 evaluation samples from the 104 in \databench for each of the five models in Table~\ref{tab:main_results}, yielding 100 samples for human review. We compute Spearman's $\rho$ and Kendall's $\tau$ between human references and two automatic judges, Gemma-4-31B and Qwen3.5-27B.
Results are reported in Table~\ref{tab:human-eval}. The Gemma-4-31B judge achieves Spearman correlations with human references of $\rho \geq 0.84$ on every metric group, peaking at $\rho = 0.898$ on the Final Score. Inter-judge agreement between the two LLM judges remains above $\rho = 0.72$ across all groups and reaches $\rho = 0.847$ on the Final Score. These results support the use of LLM-as-judge as a reproducible substitute for human evaluation during iterative model development.

\section{Related Works}\label{sec:related_works}

\paragraph{Full-duplex speech corpora.}
Public conversational corpora cover the dyadic, multi-party, and bilingual
axes only in isolation. Classical dyadic telephone corpora,
Switchboard~\cite{godfrey1992switchboard} ($\sim$260\,h),
Fisher~\cite{cieri2004fisher} ($\sim$2\,k\,h), and
HKUST/MTS~\cite{liu2006hkust} ($\sim$200\,h), provide dual-channel audio
suitable for full-duplex training but are language-monolingual and contain
only two speakers per session. Naturalistic dyadic video-chat data such as
the 850-hour CANDOR corpus~\cite{reece2023candor} adds scale and overlap
phenomena but remains English-only and dyadic. Open multi-party meeting
corpora, including AMI~\cite{carletta2006ami} ($\sim$100\,h),
ICSI~\cite{janin2003icsi} ($\sim$72\,h),
AISHELL-4~\cite{fu2021aishell4} ($\sim$120\,h), and
AliMeeting~\cite{yu2022alimeeting} ($\sim$120\,h), together amount to only
$\sim$400\,h, are each English- or Chinese-only, and were designed for
distant-microphone ASR and diarization rather than codec-frame-level
dialogue modeling. English--Chinese code-switched resources such as
ASCEND~\cite{lovenia2022ascend}, SEAME~\cite{lyu2010seame}, and
TALCS~\cite{li2022talcs} are short, single-channel, or domain-restricted to
classroom and interview settings. Reflecting
this scarcity, every recent Moshi-style system has fallen back on
proprietary in-house data or TTS-synthesized stereo dialogue that is not
publicly redistributed~\cite{kyutai2024moshi,veluri2024syncllms,zhang2025omniflatten,zeng2024glm4,huang2025stepaudio,ohashi2025jmoshi}.
The 57.6\,k-hour parallel-stream bilingual corpus released in this work
fills the long\,$\times$\,multi-party\,$\times$\,bilingual gap left by the
union of these resources.

\paragraph{Full-duplex speech dialogue models.}
End-to-end spoken dialogue systems differ in how they perceive and emit
speech, and whether they can do so simultaneously. \emph{Native full-duplex}
models maintain parallel input and output streams at codec-frame rate,
exemplified by Moshi's Mimi codec and RQ-Transformer with an Inner Monologue
text scaffold~\cite{kyutai2024moshi}, SyncLLM's periodic synchronization
tokens~\cite{veluri2024syncllms}, OmniFlatten's flattened multi-stream
sequence~\cite{zhang2025omniflatten}, and SALMONN-omni's codec-free
embedding-level formulation~\cite{yu2025salmonnomni}.
J-Moshi adapts this recipe to Japanese using 344\,h of real stereo dialogue
augmented with 602\,h of multi-stream-TTS-synthesized
data~\cite{ohashi2025jmoshi}, and FLM-Audio and RoboEgo extend native
full-duplex modeling to bilingual
English--Chinese~\cite{yao2025flmaudio,yao2025roboego}. \emph{Streaming
end-to-end} models such as GLM-4-Voice~\cite{zeng2024glm4,zeng2024scaling},
Step-Audio 1/2~\cite{huang2025stepaudio,wu2025stepaudio2},
Qwen2.5/3-Omni~\cite{xu2025qwen25omni,xu2025qwen3omni},
Kimi-Audio~\cite{kimiteam2025kimiaudio}, Baichuan-Audio/Omni~\cite{li2025baichuanaudio,li2025baichuanomni15technicalreport},
LLaMA-Omni 1/2~\cite{fang2024llamaomni,fang2025llamaomni2}, and
MinMo~\cite{chen2025minmomultimodallargelanguage} achieve low-latency interaction through
half-duplex turn-taking or time-division-multiplexed
duplexing~\citep{yao2024minicpm}. Adjacent audio-visual generation work has also explored unified autoregressive modeling of synchronized speech and video~\cite{zong2026voca}. 
Architecture search and preference-based optimization have been investigated for other language-model objectives~\cite{hu2025lmsearcher,lu2024stepcontrolleddpo,yang2025probabilityconsistentpreference}.
Across these families, none of these
systems is trained or evaluated on multi-party (${\geq}3$ speaker)
interaction, and none reports stable behavior on conversations longer than
several minutes.

\paragraph{Evaluation of long-form and multi-party speech dialogue.}
Task-specific benchmark design has been studied in multimodal mathematics, website generation, slide generation, spreadsheet understanding, and voice-assistant evaluation~\cite{wang2024mathvision,lu2025webgenbench,yang2026slidesgenbench,ren2026spreadsheetunderstanding,wang2025voiceassistanteval}.
Long-form audio benchmarks such as
BLAB~\cite{ahia2025blabbrutallylongaudio},
AudioMarathon~\cite{he2025audiomarathoncomprehensivebenchmarklongcontext},
and ChronosAudio~\cite{luo2026chronosaudio} show that audio LLMs degrade
sharply as input length grows from seconds to tens of minutes, but only evaluate
\emph{passive} listening. Spoken-dialogue benchmarks such as
VoiceBench~\cite{chen2024voicebench}, AudioBench~\cite{wang2025audiobench},
AIR-Bench~\cite{yang2024airbench}, SD-Eval~\cite{ao2024sdeval}, and
S2S-Arena~\cite{jiang2025s2sarena} probe instruction-following,
paralinguistic understanding, and arena-style preference, while Talking
Turns~\cite{arora2025talkingturns}, Full-Duplex-Bench
v1/v1.5/v2~\cite{lin2025fdb_v1,lin2025fdb_v15,lin2026fdb_v2},
MTR-DuplexBench~\cite{zhang2026mtrduplexbench},
FD-Bench~\cite{peng2025fdbench}, and MTalk-Bench~\cite{du2025mtalkbench}
target turn-taking, overlap, and multi-turn dynamics.
URO-Bench~\cite{yan2025urobench} is the only existing speech-to-speech
benchmark to combine bilingual English--Chinese evaluation with multi-round
and paralinguistic dimensions. On the multi-party side, prior evaluation has
been almost exclusively text-based, including addressee-and-response
selection~\cite{zhang2018addressee}, pre-trained representations of speaker
roles~\cite{gu2021mpcbert}, diagnostic studies of LLM behavior on
multi-party conversations~\cite{penzo2024multiparty}, and recent multi-modal
triadic addressee benchmarks~\cite{inoue2025addressee}, complemented on the
audio side by voice-activity-projection turn-taking
models~\cite{ekstedt2022vap,inoue2024vap}. None of these benchmarks
evaluates one-model-many-user multi-party
interaction longer than a few minutes in an interactive
full-duplex setting, the joint gap that
MultiTalkBench is designed to fill.
Work on visual and document reasoning further illustrates the use of task-specific representations and intermediate computation across modalities~\cite{wang2025mathcodervl,duan2026codeplotcot,shi2026mathcanvas,xiao2025adaptivemarkuplanguage}.
\section{Conclusion} \label{sec:conclusion}

In this work, we presented MultiTalk, addressing data, benchmark, and modeling gaps in open-source full-duplex speech systems for long-horizon, multi-party, bilingual interaction. We release a 57.6k-hour parallel-stream corpus (\datapt for bilingual pre-training, \dataft for multi-party fine-tuning) and \databench, the first benchmark probing long, multi-party, bilingual full-duplex dialogue jointly. Trained under our two-phase recipe, \modelname raises the overall \databench score by $+182\%$ over its Moshiko-7B initialization, unlocks Chinese from $0.36$ to $16.27$ while preserving English, and exceeds all four open-source baselines including a 30B mixture-of-experts system, while remaining well below the $68.74$ human reference.
These results suggest two readings: multi-party competence in open full-duplex models is bounded by training distribution, and the gap to human performance shows that \databench remains far from saturated. Promising directions include dialogue-aware synthesis for closer train-test acoustic match, extension to additional languages, and scaling the recipe to larger backbones.

\section{Acknowledgements}
This work is supported in part by the Centre for Perceptual and In-teractive Intelligence (CPII) Ltd under the Innovation and Technology Commission (ITC)'s InnoHK.

\bibliographystyle{plain}
\bibliography{neurips_2026}

\begin{thebibliography}{10}

\bibitem{abbo2025multiparty}
Giulio~Antonio Abbo, Maria~Jose Pinto-Bernal, Martijn Catrycke, and Tony Belpaeme.
\newblock Fast multi-party open-ended conversation with a social robot, 2025.

\bibitem{ahia2025blabbrutallylongaudio}
Orevaoghene Ahia, Martijn Bartelds, Kabir Ahuja, Hila Gonen, Valentin Hofmann, Siddhant Arora, Shuyue~Stella Li, Vishal Puttagunta, Mofetoluwa Adeyemi, Charishma Buchireddy, Ben Walls, Noah Bennett, Shinji Watanabe, Noah~A. Smith, Yulia Tsvetkov, and Sachin Kumar.
\newblock Blab: Brutally long audio bench, 2025.

\bibitem{anthropic2026sonnet46card}
{Anthropic}.
\newblock Claude sonnet 4.6 system card.
\newblock Technical report, Anthropic, February 2026.

\bibitem{ao2024sdeval}
Junyi Ao, Yuancheng Wang, Xiaohai Tian, Dekun Chen, Jun Zhang, Lu~Lu, Yuxuan Wang, Haizhou Li, and Zhizheng Wu.
\newblock Sd-eval: A benchmark dataset for spoken dialogue understanding beyond words.
\newblock {\em Advances in Neural Information Processing Systems}, 37:56898--56918, 2024.

\bibitem{arora2025talkingturns}
Siddhant Arora, Zhiyun Lu, Chung-Cheng Chiu, Ruoming Pang, and Shinji Watanabe.
\newblock Talking turns: Benchmarking audio foundation models on turn-taking dynamics, 2025.

\bibitem{carletta2006ami}
Jean Carletta, Simone Ashby, Sebastien Bourban, Mike Flynn, Ma{\"e}l Guillemot, Thomas Hain, Jaroslav Kadlec, Vasilis Karaiskos, Wessel Kraaij, Melissa Kronenthal, Guillaume Lathoud, Mike Lincoln, Agnes~Lisowska Masson, Iain McCowan, Wilfried Post, Dennis Reidsma, and Pierre~D. Wellner.
\newblock The ami meeting corpus: A pre-announcement.
\newblock In {\em Machine Learning for Multimodal Interaction}, 2005.

\bibitem{chen2025minmomultimodallargelanguage}
Qian Chen, Yafeng Chen, Yanni Chen, Mengzhe Chen, Yingda Chen, Chong Deng, Zhihao Du, Ruize Gao, Changfeng Gao, Zhifu Gao, Yabin Li, Xiang Lv, Jiaqing Liu, Haoneng Luo, Bin Ma, Chongjia Ni, Xian Shi, Jialong Tang, Hui Wang, Hao Wang, Wen Wang, Yuxuan Wang, Yunlan Xu, Fan Yu, Zhijie Yan, Yexin Yang, Baosong Yang, Xian Yang, Guanrou Yang, Tianyu Zhao, Qinglin Zhang, Shiliang Zhang, Nan Zhao, Pei Zhang, Chong Zhang, and Jinren Zhou.
\newblock Minmo: A multimodal large language model for seamless voice interaction, 2025.

\bibitem{chen2024voicebench}
Yiming Chen, Xianghu Yue, Chen Zhang, Xiaoxue Gao, Robby~T. Tan, and Haizhou Li.
\newblock Voicebench: Benchmarking llm-based voice assistants, 2024.

\bibitem{cieri2004fisher}
Christopher Cieri, David Miller, and Kevin Walker.
\newblock The fisher corpus: a resource for the next generations of speech-to-text.
\newblock In Maria~Teresa Lino, Maria~Francisca Xavier, F{\'a}tima Ferreira, Rute Costa, and Raquel Silva, editors, {\em Proceedings of the Fourth International Conference on Language Resources and Evaluation ({LREC}{'}04)}, Lisbon, Portugal, May 2004. European Language Resources Association (ELRA).

\bibitem{aishell5}
Yuhang Dai, He~Wang, Xingchen Li, Zihan Zhang, Shuiyuan Wang, Lei Xie, Xin Xu, Hongxiao Guo, Shaoji Zhang, Hui Bu, and Wei Chen.
\newblock Aishell-5: The first open-source in-car multi-channel multi-speaker speech dataset for automatic speech diarization and recognition, 2025.

\bibitem{kyutai2024moshi}
Alexandre D\'efossez, Laurent Mazar\'e, Manu Orsini, Am\'elie Royer, Patrick P\'erez, Herv\'e J\'egou, Edouard Grave, and Neil Zeghidour.
\newblock Moshi: a speech-text foundation model for real-time dialogue.
\newblock Technical report, 2024.

\bibitem{du2025mtalkbench}
Yuhao Du, Qianwei Huang, Guo Zhu, Zhanchen Dai, Shunian Chen, Qiming Zhu, Le~Pan, Minghao Chen, Yuhao Zhang, Li~Zhou, Benyou Wang, and Haizhou Li.
\newblock Mtalk-bench: Evaluating speech-to-speech models in multi-turn dialogues via arena-style and rubrics protocols, 2025.

\bibitem{duan2026codeplotcot}
Chengqi Duan, Kaiyue Sun, Rongyao Fang, Manyuan Zhang, Yan Feng, Ying Luo, Yufang Liu, Ke~Wang, Peng Pei, Xunliang Cai, Hongsheng Li, Yi~Ma, and Xihui Liu.
\newblock Codeplot-cot: Mathematical visual reasoning by thinking with code-driven images.
\newblock In {\em Proceedings of the IEEE/CVF Conference on Computer Vision and Pattern Recognition (CVPR) Findings}, pages 9586--9596, June 2026.

\bibitem{dubey2023dns}
Harishchandra Dubey, Ashkan Aazami, Vishak Gopal, Babak Naderi, Sebastian Braun, Ross Cutler, Hannes Gamper, Mehrsa Golestaneh, and Robert Aichner.
\newblock Icassp 2023 deep noise suppression challenge.
\newblock In {\em ICASSP}, 2023.

\bibitem{ekstedt2022vap}
Erik Ekstedt and Gabriel Skantze.
\newblock Voice activity projection: Self-supervised learning of turn-taking events, 2022.

\bibitem{fang2024llamaomni}
Qingkai Fang, Shoutao Guo, Yan Zhou, Zhengrui Ma, Shaolei Zhang, and Yang Feng.
\newblock Llama-omni: Seamless speech interaction with large language models.
\newblock {\em arXiv preprint arXiv:2409.06666}, 2024.

\bibitem{fang2025llamaomni2}
Qingkai Fang, Yan Zhou, Shoutao Guo, Shaolei Zhang, and Yang Feng.
\newblock Llama-omni2: Llm-based real-time spoken chatbot with autoregressive streaming speech synthesis.
\newblock {\em arXiv preprint arXiv:2505.02625}, 2025.

\bibitem{fu2021aishell4}
Yihui Fu, Luyao Cheng, Shubo Lv, Yukai Jv, Yuxiang Kong, Zhuo Chen, Yanxin Hu, Lei Xie, Jian Wu, Hui Bu, Xin Xu, Jun Du, and Jingdong Chen.
\newblock Aishell-4: An open source dataset for speech enhancement, separation, recognition and speaker diarization in conference scenario.
\newblock 2021.

\bibitem{godfrey1992switchboard}
J.J. Godfrey, E.C. Holliman, and J.~McDaniel.
\newblock Switchboard: telephone speech corpus for research and development.
\newblock In {\em [Proceedings] ICASSP-92: 1992 IEEE International Conference on Acoustics, Speech, and Signal Processing}, volume~1, pages 517--520 vol.1, 1992.

\bibitem{gemma4deepmind2026}
{Google DeepMind}.
\newblock Gemma 4: Byte for byte, the most capable open models.
\newblock \url{https://blog.google/innovation-and-ai/technology/developers-tools/gemma-4/}, 2026.

\bibitem{gu2021mpcbert}
Jia-Chen Gu, Chongyang Tao, Zhen-Hua Ling, Can Xu, Xiubo Geng, and Daxin Jiang.
\newblock Mpc-bert: A pre-trained language model for multi-party conversation understanding, 2021.

\bibitem{he2025audiomarathoncomprehensivebenchmarklongcontext}
Peize He, Zichen Wen, Yubo Wang, Yuxuan Wang, Xiaoqian Liu, Jiajie Huang, Zehui Lei, Zhuangcheng Gu, Xiangqi Jin, Jiabing Yang, Kai Li, Zhifei Liu, Weijia Li, Cunxiang Wang, Conghui He, and Linfeng Zhang.
\newblock Audiomarathon: A comprehensive benchmark for long-context audio understanding and efficiency in audio llms, 2025.

\bibitem{hu2025lmsearcher}
Yuxuan Hu, Jihao Liu, Ke~Wang, Jinliang Zheng, Weikang Shi, Manyuan Zhang, Qi~Dou, Rui Liu, Aojun Zhou, and Hongsheng Li.
\newblock {LM}-searcher: Cross-domain neural architecture search with {LLM}s via unified numerical encoding.
\newblock In Christos Christodoulopoulos, Tanmoy Chakraborty, Carolyn Rose, and Violet Peng, editors, {\em Proceedings of the 2025 Conference on Empirical Methods in Natural Language Processing}, pages 9408--9421, Suzhou, China, November 2025. Association for Computational Linguistics.

\bibitem{huang2025stepaudio}
Ailin Huang, Boyong Wu, Bruce Wang, Chao Yan, Chen Hu, Chengli Feng, Fei Tian, Feiyu Shen, Jingbei Li, Mingrui Chen, Peng Liu, Ruihang Miao, Wang You, Xi~Chen, Xuerui Yang, Yechang Huang, Yuxiang Zhang, Zheng Gong, Zixin Zhang, Hongyu Zhou, Jianjian Sun, Brian Li, Chengting Feng, Changyi Wan, Hanpeng Hu, Jianchang Wu, Jiangjie Zhen, Ranchen Ming, Song Yuan, Xuelin Zhang, Yu~Zhou, Bingxin Li, Buyun Ma, Hongyuan Wang, Kang An, Wei Ji, Wen Li, Xuan Wen, Xiangwen Kong, Yuankai Ma, Yuanwei Liang, Yun Mou, Bahtiyar Ahmidi, Bin Wang, Bo~Li, Changxin Miao, Chen Xu, Chenrun Wang, Dapeng Shi, Deshan Sun, Dingyuan Hu, Dula Sai, Enle Liu, Guanzhe Huang, Gulin Yan, Heng Wang, Haonan Jia, Haoyang Zhang, Jiahao Gong, Junjing Guo, Jiashuai Liu, Jiahong Liu, Jie Feng, Jie Wu, Jiaoren Wu, Jie Yang, Jinguo Wang, Jingyang Zhang, Junzhe Lin, Kaixiang Li, Lei Xia, Li~Zhou, Liang Zhao, Longlong Gu, Mei Chen, Menglin Wu, Ming Li, Mingxiao Li, Mingliang Li, Mingyao Liang, Na~Wang, Nie Hao, Qiling Wu, Qinyuan Tan, Ran Sun, Shuai
  Shuai, Shaoliang Pang, Shiliang Yang, Shuli Gao, Shanshan Yuan, Siqi Liu, Shihong Deng, Shilei Jiang, Sitong Liu, Tiancheng Cao, Tianyu Wang, Wenjin Deng, Wuxun Xie, Weipeng Ming, Wenqing He, Wen Sun, Xin Han, Xin Huang, Xiaomin Deng, Xiaojia Liu, Xin Wu, Xu~Zhao, Yanan Wei, Yanbo Yu, Yang Cao, Yangguang Li, Yangzhen Ma, Yanming Xu, Yaoyu Wang, Yaqiang Shi, Yilei Wang, Yizhuang Zhou, Yinmin Zhong, Yang Zhang, Yaoben Wei, Yu~Luo, Yuanwei Lu, Yuhe Yin, Yuchu Luo, Yuanhao Ding, Yuting Yan, Yaqi Dai, Yuxiang Yang, Zhe Xie, Zheng Ge, Zheng Sun, Zhewei Huang, Zhichao Chang, Zhisheng Guan, Zidong Yang, Zili Zhang, Binxing Jiao, Daxin Jiang, Heung-Yeung Shum, Jiansheng Chen, Jing Li, Shuchang Zhou, Xiangyu Zhang, Xinhao Zhang, and Yibo Zhu.
\newblock Step-audio: Unified understanding and generation in intelligent speech interaction, 2025.

\bibitem{inoue2024vap}
Koji Inoue, Bing'er Jiang, Erik Ekstedt, Tatsuya Kawahara, and Gabriel Skantze.
\newblock Multilingual turn-taking prediction using voice activity projection, 2024.

\bibitem{inoue2025addressee}
Koji Inoue, Divesh Lala, Mikey Elmers, Keiko Ochi, and Tatsuya Kawahara.
\newblock An llm benchmark for addressee recognition in multi-modal multi-party dialogue, 2025.

\bibitem{janin2003icsi}
A.~Janin, D.~Baron, J.~Edwards, D.~Ellis, D.~Gelbart, N.~Morgan, B.~Peskin, T.~Pfau, E.~Shriberg, A.~Stolcke, and C.~Wooters.
\newblock The icsi meeting corpus.
\newblock In {\em 2003 IEEE International Conference on Acoustics, Speech, and Signal Processing, 2003. Proceedings. (ICASSP '03).}, volume~1, pages I--I, 2003.

\bibitem{jiang2025s2sarena}
Feng Jiang, Zhiyu Lin, Fan Bu, Yuhao Du, Benyou Wang, and Haizhou Li.
\newblock S2s-arena, evaluating speech2speech protocols on instruction following with paralinguistic information, 2025.

\bibitem{kimiteam2025kimiaudio}
KimiTeam, Ding Ding, Zeqian Ju, Yichong Leng, Songxiang Liu, Tong Liu, Zeyu Shang, Kai Shen, Wei Song, Xu~Tan, Heyi Tang, Zhengtao Wang, Chu Wei, Yifei Xin, Xinran Xu, Jianwei Yu, Yutao Zhang, Xinyu Zhou, Y.~Charles, Jun Chen, Yanru Chen, Yulun Du, Weiran He, Zhenxing Hu, Guokun Lai, Qingcheng Li, Yangyang Liu, Weidong Sun, Jianzhou Wang, Yuzhi Wang, Yuefeng Wu, Yuxin Wu, Dongchao Yang, Hao Yang, Ying Yang, Zhilin Yang, Aoxiong Yin, Ruibin Yuan, Yutong Zhang, and Zaida Zhou.
\newblock Kimi-audio technical report, 2025.

\bibitem{li2022talcs}
Chengfei Li, Shuhao Deng, Yaoping Wang, Guangjing Wang, Yaguang Gong, Changbin Chen, and Jinfeng Bai.
\newblock Talcs: An open-source mandarin-english code-switching corpus and a speech recognition baseline.
\newblock 2022.

\bibitem{li2025baichuanaudio}
Tianpeng Li, Jun Liu, Tao Zhang, Yuanbo Fang, Da~Pan, Mingrui Wang, Zheng Liang, Zehuan Li, Mingan Lin, Guosheng Dong, Jianhua Xu, Haoze Sun, Zenan Zhou, and Weipeng Chen.
\newblock Baichuan-audio: A unified framework for end-to-end speech interaction, 2025.

\bibitem{li2025baichuanomni15technicalreport}
Yadong Li, Jun Liu, Tao Zhang, Tao Zhang, Song Chen, Tianpeng Li, Zehuan Li, Lijun Liu, Lingfeng Ming, Guosheng Dong, Da~Pan, Chong Li, Yuanbo Fang, Dongdong Kuang, Mingrui Wang, Chenglin Zhu, Youwei Zhang, Hongyu Guo, Fengyu Zhang, Yuran Wang, Bowen Ding, Wei Song, Xu~Li, Yuqi Huo, Zheng Liang, Shusen Zhang, Xin Wu, Shuai Zhao, Linchu Xiong, Yozhen Wu, Jiahui Ye, Wenhao Lu, Bowen Li, Yan Zhang, Yaqi Zhou, Xin Chen, Lei Su, Hongda Zhang, Fuzhong Chen, Xuezhen Dong, Na~Nie, Zhiying Wu, Bin Xiao, Ting Li, Shunya Dang, Ping Zhang, Yijia Sun, Jincheng Wu, Jinjie Yang, Xionghai Lin, Zhi Ma, Kegeng Wu, Jia li, Aiyuan Yang, Hui Liu, Jianqiang Zhang, Xiaoxi Chen, Guangwei Ai, Wentao Zhang, Yicong Chen, Xiaoqin Huang, Kun Li, Wenjing Luo, Yifei Duan, Lingling Zhu, Ran Xiao, Zhe Su, Jiani Pu, Dian Wang, Xu~Jia, Tianyu Zhang, Mengyu Ai, Mang Wang, Yujing Qiao, Lei Zhang, Yanjun Shen, Fan Yang, Miao Zhen, Yijie Zhou, Mingyang Chen, Fei Li, Chenzheng Zhu, Keer Lu, Yaqi Zhao, Hao Liang, Youquan Li, Yanzhao Qin, Linzhuang
  Sun, Jianhua Xu, Haoze Sun, Mingan Lin, Zenan Zhou, and Weipeng Chen.
\newblock Baichuan-omni-1.5 technical report, 2025.

\bibitem{lin2026fdb_v2}
Guan-Ting Lin, Shih-Yun~Shan Kuan, Jiatong Shi, Kai-Wei Chang, Siddhant Arora, Shinji Watanabe, and Hung-yi Lee.
\newblock Full-duplex-bench-v2: A multi-turn evaluation framework for duplex dialogue systems with an automated examiner.
\newblock {\em arXiv preprint arXiv:2510.07838}, 2026.

\bibitem{lin2025fdb_v15}
Guan-Ting Lin, Shih-Yun~Shan Kuan, Qirui Wang, Jiachen Lian, Tingle Li, and Hung-yi Lee.
\newblock Full-duplex-bench v1. 5: Evaluating overlap handling for full-duplex speech models.
\newblock {\em arXiv preprint arXiv:2507.23159}, 2025.

\bibitem{lin2025fdb_v1}
Guan-Ting Lin, Jiachen Lian, Tingle Li, Qirui Wang, Gopala Anumanchipalli, Alexander~H Liu, and Hung-yi Lee.
\newblock Full-duplex-bench: A benchmark to evaluate full-duplex spoken dialogue models on turn-taking capabilities.
\newblock {\em arXiv preprint arXiv:2503.04721}, 2025.

\bibitem{liu2006hkust}
Yi~Liu, Pascale Fung, Yongsheng Yang, Christopher Cieri, Shudong Huang, and David Graff.
\newblock Hkust/mts: A very large scale mandarin telephone speech corpus.
\newblock In Qiang Huo, Bin Ma, Eng-Siong Chng, and Haizhou Li, editors, {\em Chinese Spoken Language Processing}, pages 724--735, Berlin, Heidelberg, 2006. Springer Berlin Heidelberg.

\bibitem{lovenia2022ascend}
Holy Lovenia, Samuel Cahyawijaya, Genta Winata, Peng Xu, Yan Xu, Zihan Liu, Rita Frieske, Tiezheng Yu, Wenliang Dai, Elham~J. Barezi, Qifeng Chen, Xiaojuan Ma, Bertram Shi, and Pascale Fung.
\newblock {ASCEND}: A spontaneous {C}hinese-{E}nglish dataset for code-switching in multi-turn conversation.
\newblock In Nicoletta Calzolari, Fr{\'e}d{\'e}ric B{\'e}chet, Philippe Blache, Khalid Choukri, Christopher Cieri, Thierry Declerck, Sara Goggi, Hitoshi Isahara, Bente Maegaard, Joseph Mariani, H{\'e}l{\`e}ne Mazo, Jan Odijk, and Stelios Piperidis, editors, {\em Proceedings of the Thirteenth Language Resources and Evaluation Conference}, pages 7259--7268, Marseille, France, June 2022. European Language Resources Association.

\bibitem{lu2026webgenagent}
Zimu Lu, Houxing Ren, Yunqiao Yang, Ke~Wang, Zhuofan Zong, Junting Pan, Mingjie Zhan, and Hongsheng Li.
\newblock Webgen-agent: Enhancing interactive website generation with multi-level feedback and step-level reinforcement learning.
\newblock In C.~Vondrick, B.~Hariharan, C.~Raffel, L.~Pinto, D.~Yang, and A.~Faust, editors, {\em International Conference on Learning Representations}, volume 2026, pages 83478--83525, 2026.

\bibitem{lu2026fullstackagent}
Zimu Lu, Houxing Ren, Yunqiao Yang, Ke~Wang, Zhuofan Zong, Mingjie Zhan, and Hongsheng Li.
\newblock Fullstack-agent: Enhancing agentic full-stack web coding via development-oriented testing and repository back-translation, 2026.

\bibitem{lu2025webgenbench}
Zimu Lu, Yunqiao Yang, Houxing Ren, Haotian Hou, Han Xiao, Ke~Wang, Weikang Shi, Aojun Zhou, Mingjie Zhan, and Hongsheng Li.
\newblock Webgen-bench: Evaluating llms on generating interactive and functional websites from scratch.
\newblock In D.~Belgrave, C.~Zhang, H.~Lin, R.~Pascanu, P.~Koniusz, M.~Ghassemi, and N.~Chen, editors, {\em Advances in Neural Information Processing Systems}, volume 38, Main Conference. Curran Associates, Inc., 2025.

\bibitem{lu2024mathgenie}
Zimu Lu, Aojun Zhou, Houxing Ren, Ke~Wang, Weikang Shi, Junting Pan, Mingjie Zhan, and Hongsheng Li.
\newblock {M}ath{G}enie: Generating synthetic data with question back-translation for enhancing mathematical reasoning of {LLM}s.
\newblock In Lun-Wei Ku, Andre Martins, and Vivek Srikumar, editors, {\em Proceedings of the 62nd Annual Meeting of the Association for Computational Linguistics (Volume 1: Long Papers)}, pages 2732--2747, Bangkok, Thailand, August 2024. Association for Computational Linguistics.

\bibitem{lu2024stepcontrolleddpo}
Zimu Lu, Aojun Zhou, Ke~Wang, Houxing Ren, Weikang Shi, Junting Pan, Mingjie Zhan, and Hongsheng Li.
\newblock Step-controlled dpo: Leveraging stepwise error for enhanced mathematical reasoning, 2024.

\bibitem{lu2025mathcoder2}
Zimu Lu, Aojun Zhou, Ke~Wang, Houxing Ren, Weikang Shi, Junting Pan, Mingjie Zhan, and Hongsheng Li.
\newblock Mathcoder2: Better math reasoning from continued pretraining on model-translated mathematical code.
\newblock In Y.~Yue, A.~Garg, N.~Peng, F.~Sha, and R.~Yu, editors, {\em International Conference on Learning Representations}, volume 2025, pages 76545--76565, 2025.

\bibitem{luo2026chronosaudio}
Kaiwen Luo, Liang Lin, Yibo Zhang, Moayad Aloqaily, Dexian Wang, Zhenhong Zhou, Junwei Zhang, Kun Wang, Li~Sun, and Qingsong Wen.
\newblock Chronosaudio: A comprehensive long-audio benchmark for evaluating audio-large language models, 2026.

\bibitem{lyu2010seame}
Dau-Cheng Lyu, Tien~Ping Tan, Chng~Eng Siong, and Haizhou Li.
\newblock Seame: a mandarin-english code-switching speech corpus in south-east asia.
\newblock In {\em Interspeech}, 2010.

\bibitem{mcauliffe2017mfa}
Michael McAuliffe, Michaela Socolof, Sarah Mihuc, Michael Wagner, and Morgan Sonderegger.
\newblock {Montreal Forced Aligner: Trainable Text-Speech Alignment Using Kaldi}.
\newblock In {\em {Interspeech 2017}}, pages 498--502, 2017.

\bibitem{ohashi2025jmoshi}
Atsumoto Ohashi, Shinya Iizuka, Jingjing Jiang, and Ryuichiro Higashinaka.
\newblock Towards a japanese full-duplex spoken dialogue system.
\newblock In {\em Proceedings of the 26th Interspeech Conference}, 2025.

\bibitem{openai2024gpt4ocard}
OpenAI.
\newblock Gpt-4o system card, 2024.

\bibitem{peng2025fdbench}
Yizhou Peng, Yi-Wen Chao, Dianwen Ng, Yukun Ma, Chongjia Ni, Bin Ma, and Eng~Siong Chng.
\newblock Fd-bench: A full-duplex benchmarking pipeline designed for full duplex spoken dialogue systems, 2025.

\bibitem{penzo2024multiparty}
Nicol{\`o} Penzo, Maryam Sajedinia, Bruno Lepri, Sara Tonelli, and Marco Guerini.
\newblock Do {LLM}s suffer from multi-party hangover? a diagnostic approach to addressee recognition and response selection in conversations.
\newblock In Yaser Al-Onaizan, Mohit Bansal, and Yun-Nung Chen, editors, {\em Proceedings of the 2024 Conference on Empirical Methods in Natural Language Processing}, pages 11210--11233, Miami, Florida, USA, November 2024. Association for Computational Linguistics.

\bibitem{qwen35}
{Qwen Team}.
\newblock {Qwen3.5}: Towards native multimodal agents, February 2026.

\bibitem{reece2023candor}
Andrew Reece, Gus Cooney, Peter Bull, Christine Chung, Bryn Dawson, Casey Fitzpatrick, Tamara Glazer, Dean Knox, Alex Liebscher, and Sebastian Marin.
\newblock Advancing an interdisciplinary science of conversation: Insights from a large multimodal corpus of human speech, 2022.

\bibitem{ren2025fillinthemiddle}
Houxing Ren, Zimu Lu, Weikang Shi, Haotian Hou, Yunqiao Yang, Ke~Wang, Aojun Zhou, Junting Pan, Mingjie Zhan, and Hongsheng Li.
\newblock Alignment with fill-in-the-middle for enhancing code generation.
\newblock In Christos Christodoulopoulos, Tanmoy Chakraborty, Carolyn Rose, and Violet Peng, editors, {\em Proceedings of the 2025 Conference on Empirical Methods in Natural Language Processing}, pages 8304--8320, Suzhou, China, November 2025. Association for Computational Linguistics.

\bibitem{ren2026spreadsheetunderstanding}
Houxing Ren, Mingjie Zhan, Zimu Lu, Ke~Wang, Yunqiao Yang, Haotian Hou, and Hongsheng Li.
\newblock Towards robust real-world spreadsheet understanding with multi-agent multi-format reasoning.
\newblock In Maria Liakata, Viviane~P. Moreira, Jiajun Zhang, and David Jurgens, editors, {\em Proceedings of the 64th Annual Meeting of the {A}ssociation for {C}omputational {L}inguistics (Volume 1: Long Papers)}, pages 1906--1933, San Diego, California, United States, July 2026. Association for Computational Linguistics.

\bibitem{ren2026editbasedrefinement}
Houxing Ren, Mingjie Zhan, Zimu Lu, Ke~Wang, Yunqiao Yang, Haotian Hou, Junting Pan, and Hongsheng Li.
\newblock Edit-based refinement for parallel masked diffusion language models, 2026.

\bibitem{ren2024characterleveltextinfilling}
Houxing Ren, Mingjie Zhan, Zhongyuan Wu, and Hongsheng Li.
\newblock Empowering character-level text infilling by eliminating sub-tokens, 2024.

\bibitem{ren2025reflectioncoder}
Houxing Ren, Mingjie Zhan, Zhongyuan Wu, Aojun Zhou, Junting Pan, and Hongsheng Li.
\newblock {R}eflection{C}oder: Learning from reflection sequence for enhanced one-off code generation.
\newblock In Wanxiang Che, Joyce Nabende, Ekaterina Shutova, and Mohammad~Taher Pilehvar, editors, {\em Proceedings of the 63rd Annual Meeting of the Association for Computational Linguistics (Volume 1: Long Papers)}, pages 9999--10020, Vienna, Austria, July 2025. Association for Computational Linguistics.

\bibitem{nvidia2026personaplex}
Rajarshi Roy, Jonathan Raiman, Sang gil Lee, Teodor-Dumitru Ene, Robert Kirby, Sungwon Kim, Jaehyeon Kim, and Bryan Catanzaro.
\newblock Personaplex: Voice and role control for full duplex conversational speech models, 2026.

\bibitem{saravia2018carer}
Elvis Saravia, Hsien-Chi~Toby Liu, Yen-Hao Huang, Junlin Wu, and Yi-Shin Chen.
\newblock {CARER}: Contextualized affect representations for emotion recognition.
\newblock In {\em Proceedings of the 2018 Conference on Empirical Methods in Natural Language Processing}, pages 3687--3697, Brussels, Belgium, October-November 2018. Association for Computational Linguistics.

\bibitem{shi2026mathcanvas}
Weikang Shi, Aldrich Yu, Rongyao Fang, Houxing Ren, Ke~Wang, Aojun Zhou, Changyao Tian, Xinyu Fu, Yuxuan Hu, Zimu Lu, Linjiang Huang, Si~Liu, Rui Liu, and Hongsheng Li.
\newblock {M}ath{C}anvas: Intrinsic visual chain-of-thought for multimodal mathematical reasoning.
\newblock In Maria Liakata, Viviane~P. Moreira, Jiajun Zhang, and David Jurgens, editors, {\em Proceedings of the 64th Annual Meeting of the {A}ssociation for {C}omputational {L}inguistics (Volume 1: Long Papers)}, pages 27933--27954, San Diego, California, United States, July 2026. Association for Computational Linguistics.

\bibitem{skantze2021turntaking}
Gabriel Skantze.
\newblock Turn-taking in conversational systems and human-robot interaction: A review.
\newblock {\em Comput. Speech Lang.}, 67:101178, 2021.

\bibitem{geminiteam2025geminifamilyhighlycapable}
Gemini Team, Rohan Anil, Sebastian Borgeaud, Jean-Baptiste Alayrac, Jiahui Yu, Radu Soricut, Johan Schalkwyk, Andrew~M. Dai, Anja Hauth, Katie Millican, David Silver, Melvin Johnson, Ioannis Antonoglou, Julian Schrittwieser, Amelia Glaese, Jilin Chen, Emily Pitler, Timothy Lillicrap, Angeliki Lazaridou, Orhan Firat, James Molloy, Michael Isard, Paul~R. Barham, Tom Hennigan, Benjamin Lee, Fabio Viola, Malcolm Reynolds, Yuanzhong Xu, Ryan Doherty, Eli Collins, Clemens Meyer, Eliza Rutherford, Erica Moreira, Kareem Ayoub, Megha Goel, Jack Krawczyk, Cosmo Du, Ed~Chi, Heng-Tze Cheng, Eric Ni, Purvi Shah, Patrick Kane, Betty Chan, Manaal Faruqui, Aliaksei Severyn, Hanzhao Lin, YaGuang Li, Yong Cheng, Abe Ittycheriah, Mahdis Mahdieh, Mia Chen, Pei Sun, Dustin Tran, Sumit Bagri, Balaji Lakshminarayanan, Jeremiah Liu, Andras Orban, Fabian Güra, Hao Zhou, Xinying Song, Aurelien Boffy, Harish Ganapathy, Steven Zheng, HyunJeong Choe, Ágoston Weisz, Tao Zhu, Yifeng Lu, Siddharth Gopal, Jarrod Kahn, Maciej Kula, Jeff
  Pitman, Rushin Shah, Emanuel Taropa, Majd~Al Merey, Martin Baeuml, Zhifeng Chen, Laurent~El Shafey, Yujing Zhang, Olcan Sercinoglu, George Tucker, et~al.
\newblock Gemini: A family of highly capable multimodal models, 2025.

\bibitem{comanici2025gemini25pushingfrontier}
Gemini~2.5 Team.
\newblock Gemini 2.5: Pushing the frontier with advanced reasoning, multimodality, long context, and next generation agentic capabilities, 2025.

\bibitem{tian2025amdistillation}
Xiaoyu Tian, Yunjie Ji, Haotian Wang, Shuaiting Chen, Sitong Zhao, Yiping Peng, Han Zhao, and Xiangang Li.
\newblock Not all correct answers are equal: Why your distillation source matters, 2025.

\bibitem{veluri2024syncllms}
Bandhav Veluri, Benjamin~N Peloquin, Bokai Yu, Hongyu Gong, and Shyamnath Gollakota.
\newblock Beyond turn-based interfaces: Synchronous llms as full-duplex dialogue agents, 2024.

\bibitem{wang2025audiobench}
Bin Wang, Xunlong Zou, Geyu Lin, Shuo Sun, Zhuohan Liu, Wenyu Zhang, Zhengyuan Liu, AiTi Aw, and Nancy~F. Chen.
\newblock Audiobench: A universal benchmark for audio large language models, 2025.

\bibitem{wang2024mathvision}
Ke~Wang, Junting Pan, Weikang Shi, Zimu Lu, Houxing Ren, Aojun Zhou, Mingjie Zhan, and Hongsheng Li.
\newblock Measuring multimodal mathematical reasoning with math-vision dataset.
\newblock In {\em The Thirty-eight Conference on Neural Information Processing Systems Datasets and Benchmarks Track}, 2024.

\bibitem{wang2025mathcodervl}
Ke~Wang, Junting Pan, Linda Wei, Aojun Zhou, Weikang Shi, Zimu Lu, Han Xiao, Yunqiao Yang, Houxing Ren, Mingjie Zhan, and Hongsheng Li.
\newblock {M}ath{C}oder-{VL}: Bridging vision and code for enhanced multimodal mathematical reasoning.
\newblock In Wanxiang Che, Joyce Nabende, Ekaterina Shutova, and Mohammad~Taher Pilehvar, editors, {\em Findings of the Association for Computational Linguistics: ACL 2025}, pages 2505--2534, Vienna, Austria, July 2025. Association for Computational Linguistics.

\bibitem{wang2025voiceassistanteval}
Ke~Wang, Houxing Ren, Zimu Lu, Mingjie Zhan, and Hongsheng Li.
\newblock Voiceassistant-eval: Benchmarking ai assistants across listening, speaking, and viewing, 2025.

\bibitem{wang2024mathcoder}
Ke~Wang, Houxing Ren, Aojun Zhou, Zimu Lu, Sichun Luo, Weikang Shi, Renrui Zhang, Linqi Song, Mingjie Zhan, and Hongsheng Li.
\newblock Mathcoder: Seamless code integration in llms for enhanced mathematical reasoning.
\newblock In B.~Kim, Y.~Yue, S.~Chaudhuri, K.~Fragkiadaki, M.~Khan, and Y.~Sun, editors, {\em International Conference on Learning Representations}, volume 2024, pages 5009--5042, 2024.

\bibitem{watanabe2020chime6}
Shinji Watanabe, Michael Mandel, Jon Barker, Emmanuel Vincent, Ashish Arora, Xuankai Chang, Sanjeev Khudanpur, Vimal Manohar, Daniel Povey, Desh Raj, David Snyder, Aswin~Shanmugam Subramanian, Jan Trmal, Bar~Ben Yair, Christoph Boeddeker, Zhaoheng Ni, Yusuke Fujita, Shota Horiguchi, Naoyuki Kanda, Takuya Yoshioka, and Neville Ryant.
\newblock Chime-6 challenge:tackling multispeaker speech recognition for unsegmented recordings, 2020.

\bibitem{wu2025stepaudio2}
Boyong Wu, Chao Yan, Chen Hu, Cheng Yi, Chengli Feng, Fei Tian, Feiyu Shen, Gang Yu, Haoyang Zhang, Jingbei Li, Mingrui Chen, Peng Liu, Wang You, Xiangyu~Tony Zhang, Xingyuan Li, Xuerui Yang, Yayue Deng, Yechang Huang, Yuxin Li, Yuxin Zhang, Zhao You, Brian Li, Changyi Wan, Hanpeng Hu, Jiangjie Zhen, Siyu Chen, Song Yuan, Xuelin Zhang, Yimin Jiang, Yu~Zhou, Yuxiang Yang, Bingxin Li, Buyun Ma, Changhe Song, Dongqing Pang, Guoqiang Hu, Haiyang Sun, Kang An, Na~Wang, Shuli Gao, Wei Ji, Wen Li, Wen Sun, Xuan Wen, Yong Ren, Yuankai Ma, Yufan Lu, Bin Wang, Bo~Li, Changxin Miao, Che Liu, Chen Xu, Dapeng Shi, Dingyuan Hu, Donghang Wu, Enle Liu, Guanzhe Huang, Gulin Yan, Han Zhang, Hao Nie, Haonan Jia, Hongyu Zhou, Jianjian Sun, Jiaoren Wu, Jie Wu, Jie Yang, Jin Yang, Junzhe Lin, Kaixiang Li, Lei Yang, Liying Shi, Li~Zhou, Longlong Gu, Ming Li, Mingliang Li, Mingxiao Li, Nan Wu, Qi~Han, Qinyuan Tan, Shaoliang Pang, Shengjie Fan, Siqi Liu, Tiancheng Cao, Wanying Lu, Wenqing He, Wuxun Xie, Xu~Zhao, Xueqi Li, Yanbo Yu,
  Yang Yang, Yi~Liu, Yifan Lu, Yilei Wang, Yuanhao Ding, Yuanwei Liang, Yuanwei Lu, Yuchu Luo, Yuhe Yin, Yumeng Zhan, Yuxiang Zhang, Zidong Yang, Zixin Zhang, Binxing Jiao, Daxin Jiang, Heung-Yeung Shum, Jiansheng Chen, Jing Li, Xiangyu Zhang, and Yibo Zhu.
\newblock Step-audio 2 technical report, 2025.

\bibitem{xiao2025adaptivemarkuplanguage}
Han Xiao, Yina Xie, Guanxin Tan, Yinghao Chen, Rui Hu, Ke~Wang, Aojun Zhou, Hao Li, Hao Shao, Xudong Lu, Peng Gao, Yafei Wen, Xiaoxin Chen, Shuai Ren, and Hongsheng Li.
\newblock Adaptive markup language generation for contextually-grounded visual document understanding.
\newblock In {\em 2025 IEEE/CVF Conference on Computer Vision and Pattern Recognition (CVPR)}, pages 29558--29568, 2025.

\bibitem{xu2025qwen25omni}
Jin Xu, Zhifang Guo, Jinzheng He, Hangrui Hu, Ting He, Shuai Bai, Keqin Chen, Jialin Wang, Yang Fan, Kai Dang, Bin Zhang, Xiong Wang, Yunfei Chu, and Junyang Lin.
\newblock Qwen2.5-omni technical report.
\newblock {\em arXiv preprint arXiv:2503.20215}, 2025.

\bibitem{xu2025qwen3omni}
Jin Xu, Zhifang Guo, Hangrui Hu, Yunfei Chu, Xiong Wang, Jinzheng He, Yuxuan Wang, Xian Shi, Ting He, Xinfa Zhu, Yuanjun Lv, Yongqi Wang, Dake Guo, He~Wang, Linhan Ma, Pei Zhang, Xinyu Zhang, Hongkun Hao, Zishan Guo, Baosong Yang, Bin Zhang, Ziyang Ma, Xipin Wei, Shuai Bai, Keqin Chen, Xuejing Liu, Peng Wang, Mingkun Yang, Dayiheng Liu, Xingzhang Ren, Bo~Zheng, Rui Men, Fan Zhou, Bowen Yu, Jianxin Yang, Le~Yu, Jingren Zhou, and Junyang Lin.
\newblock Qwen3-omni technical report, 2025.

\bibitem{yan2025urobench}
Ruiqi Yan, Xiquan Li, Wenxi Chen, Zhikang Niu, Chen Yang, Ziyang Ma, Kai Yu, and Xie Chen.
\newblock Uro-bench: Towards comprehensive evaluation for end-to-end spoken dialogue models, 2025.

\bibitem{yang2024airbench}
Qian Yang, Jin Xu, Wenrui Liu, Yunfei Chu, Ziyue Jiang, Xiaohuan Zhou, Yichong Leng, Yuanjun Lv, Zhou Zhao, Chang Zhou, and Jingren Zhou.
\newblock Air-bench: Benchmarking large audio-language models via generative comprehension, 2024.

\bibitem{yang2026slidesgenbench}
Yunqiao Yang, Wenbo Li, Houxing Ren, Zimu Lu, Ke~Wang, Zhiyuan Huang, Zhuofan Zong, Mingjie Zhan, and Hongsheng Li.
\newblock Slidesgen-bench: Evaluating slides generation via computational and quantitative metrics, 2026.

\bibitem{yang2025probabilityconsistentpreference}
Yunqiao Yang, Houxing Ren, Zimu Lu, Ke~Wang, Weikang Shi, Aojun Zhou, Junting Pan, Mingjie Zhan, and Hongsheng Li.
\newblock Probability-consistent preference optimization for enhanced {LLM} reasoning.
\newblock In Wanxiang Che, Joyce Nabende, Ekaterina Shutova, and Mohammad~Taher Pilehvar, editors, {\em Findings of the Association for Computational Linguistics: ACL 2025}, pages 6435--6448, Vienna, Austria, July 2025. Association for Computational Linguistics.

\bibitem{yang2022magicdata}
Zehui Yang, Yifan Chen, Lei Luo, Runyan Yang, Lingxuan Ye, Gaofeng Cheng, Ji~Xu, Yaohui Jin, Qingqing Zhang, Pengyuan Zhang, Lei Xie, and Yonghong Yan.
\newblock Open source magicdata-ramc: A rich annotated mandarin conversational(ramc) speech dataset, 2022.

\bibitem{yao2025flmaudio}
Yiqun Yao, Xiang Li, Xin Jiang, Xuezhi Fang, Naitong Yu, Wenjia Ma, Aixin Sun, and Yequan Wang.
\newblock Flm-audio: Natural monologues improves native full-duplex chatbots via dual training, 2026.

\bibitem{yao2025roboego}
Yiqun Yao, Xiang Li, Xin Jiang, Xuezhi Fang, Naitong Yu, Aixin Sun, and Yequan Wang.
\newblock Roboego system card: An omnimodal model with native full duplexity, 2025.

\bibitem{yao2024minicpm}
Yuan Yao, Tianyu Yu, Ao~Zhang, Chongyi Wang, Junbo Cui, Hongji Zhu, Tianchi Cai, Haoyu Li, Weilin Zhao, Zhihui He, et~al.
\newblock Minicpm-v: A gpt-4v level mllm on your phone.
\newblock {\em arXiv preprint arXiv:2408.01800}, 2024.

\bibitem{yu2022alimeeting}
Fan Yu, Shiliang Zhang, Yihui Fu, Lei Xie, Siqi Zheng, Zhihao Du, Weilong Huang, Pengcheng Guo, Zhijie Yan, Bin Ma, Xin Xu, and Hui Bu.
\newblock M2met: The icassp 2022 multi-channel multi-party meeting transcription challenge.
\newblock 2022.

\bibitem{yu2025salmonnomni}
Wenyi Yu, Siyin Wang, Xiaoyu Yang, Xianzhao Chen, Xiaohai Tian, Jun Zhang, Guangzhi Sun, Lu~Lu, Yuxuan Wang, and Chao Zhang.
\newblock Salmonn-omni: A standalone speech llm without codec injection for full-duplex conversation, 2025.

\bibitem{zeng2024glm4}
Aohan Zeng, Zhengxiao Du, Mingdao Liu, Kedong Wang, Shengmin Jiang, Lei Zhao, Yuxiao Dong, and Jie Tang.
\newblock Glm-4-voice: Towards intelligent and human-like end-to-end spoken chatbot, 2024.

\bibitem{zeng2024scaling}
Aohan Zeng, Zhengxiao Du, Mingdao Liu, Lei Zhang, Shengmin Jiang, Yuxiao Dong, and Jie Tang.
\newblock Scaling speech-text pre-training with synthetic interleaved data, 2024.

\bibitem{zhang2026mtrduplexbench}
He~Zhang, Wenqian Cui, Haoning Xu, Xiaohui Li, Lei Zhu, Haoli Bai, Shaohua Ma, and Irwin King.
\newblock Mtr-duplexbench: Towards a comprehensive evaluation of multi-round conversations for full-duplex speech language models, 2026.

\bibitem{zhang2025omniflatten}
Qinglin Zhang, Luyao Cheng, Chong Deng, Qian Chen, Wen Wang, Siqi Zheng, Jiaqing Liu, Hai Yu, Chaohong Tan, Zhihao Du, and Shiliang Zhang.
\newblock Omniflatten: An end-to-end gpt model for seamless voice conversation, 2025.

\bibitem{zhang2018addressee}
Rui Zhang, Honglak Lee, Lazaros Polymenakos, and Dragomir Radev.
\newblock Addressee and response selection in multi-party conversations with speaker interaction rnns, 2017.

\bibitem{zhou2024gpt4codeinterpreter}
Aojun Zhou, Ke~Wang, Zimu Lu, Weikang Shi, Sichun Luo, Zipeng Qin, Shaoqing Lu, Anya Jia, Linqi Song, Mingjie Zhan, and Hongsheng Li.
\newblock Solving challenging math word problems using gpt-4 code interpreter with code-based self-verification.
\newblock In B.~Kim, Y.~Yue, S.~Chaudhuri, K.~Fragkiadaki, M.~Khan, and Y.~Sun, editors, {\em International Conference on Learning Representations}, volume 2024, pages 4468--4494, 2024.

\bibitem{zhou2025indextts2}
Siyi Zhou, Yiquan Zhou, Yi~He, Xun Zhou, Jinchao Wang, Wei Deng, and Jingchen Shu.
\newblock Indextts2: A breakthrough in emotionally expressive and duration-controlled auto-regressive zero-shot text-to-speech.
\newblock {\em arXiv preprint arXiv:2506.21619}, 2025.

\bibitem{zong2026voca}
Zhuofan Zong, Jiale Yuan, Yufei Liu, Dongzhi Jiang, Hao Shao, Zimu Lu, Ke~Wang, Yunqiao Yang, Mingjie Zhan, and Hongsheng Li.
\newblock Voca: Unified autoregressive modeling for talking audio-video generation.
\newblock In Paolo Favaro, Zuzana Kukelova, Atsuto Maki, Anna Rohrbach, Konrad Schindler, and Federico Tombari, editors, {\em Computer Vision -- ECCV 2026}, pages 138--156, Cham, 2026. Springer Nature Switzerland.

\bibitem{züfle2026factor}
Maike Züfle, Ondrej Klejch, Nicholas Sanders, Jan Niehues, Alexandra Birch, and Tsz~Kin Lam.
\newblock F-actor: Controllable conversational behaviour in full-duplex models, 2026.

\end{thebibliography}

\appendix
\section*{Appendix}

\section{Limitations}
\label{app:limitations}

\paragraph{Absolute capability headroom.}
The strongest model in our experiments, Moshi-MTB, attains a Final score of $13.15$ versus a human reference of $68.74$ on \databench\ (Table~\ref{tab:main_results}). This gap reflects two compounding factors. First, long-horizon multi-party bilingual dialogue lies beyond the reach of all open-source full-duplex systems we evaluated, so the empirical ceiling our recipe approaches is itself low. Second, our 7B-parameter backbone bounds the headroom that data alone can recover, and we have not evaluated this recipe on larger backbones. \databench\ is therefore best read as a forward-looking benchmark with substantial headroom rather than a saturated one.

\paragraph{Synthetic training audio.} \datapt\ and \dataft\ are rendered with a zero-shot TTS system (IndexTTS2) and assembled into multi-channel streams under scripted overlap and gap-compression rules. This pipeline delivers codec-frame-level alignment and controllable conversational dynamics at the 57.6k-hour scale required for full-duplex pre-training, while maintaining acoustic realism through far-field DNS noise mixing on the user channel during training (Section~\ref{sec:experiments}). We emphasize that, since \datapt/\dataft\ are synthetic while \databench\ is not, training on \datapt/\dataft\ and evaluating on \databench\ is by construction a cross-domain assessment. The strong \databench\ scores reported in Section~\ref{sec:experiments} indicate that our pipeline closes the synthetic-to-real gap to a useful extent. Extending the engine with dialogue-aware TTS or real-prosody grafting is a natural direction for future work, and would further enrich the interactional nuance available to the model.

\paragraph{Language coverage.}
The reported benchmark evaluates English and Chinese separately; it does not
test intra-sentential code-switching. Extension to lower-resource languages
may be limited by TTS quality and voice diversity, script fluency and cultural
appropriateness, language-dependent codec and ASR errors, and the availability
of real multi-party recordings and native-speaker validation of the judge.
Our results should not be assumed to generalize beyond English and Chinese.

\section{Compute Resources}
\label{app:compute}

Both phases of Moshi-MTB were trained on NVIDIA {H800-80GB} GPUs, with bf16 mixed precision. Phase 1 used approximately {560} GPU-hours. Phase 2 used approximately {33} GPU-hours.
TTS rendering with IndexTTS2 ran on {64} GPUs over {30} days for the 57.6k-hour corpus.

\section{Broader Impacts}
\label{app:broader_impacts}

\paragraph{Positive impacts.} Robust long-horizon multi-party speech models can improve accessibility (real-time captioning and turn-taking support for hearing-impaired users in group settings), education (automated facilitators in group lessons), and assistive robotics (social robots that handle reception, family, or care scenarios involving multiple humans). Releasing an open data engine and corpus also lowers the entry barrier for academic research, which has so far been blocked by the proprietary in-house data used by every Moshi-style system.

\paragraph{Negative impacts and mitigations.} The same capabilities enable risks. First, voice impersonation: improved zero-shot multi-speaker generation could be misused for deepfake calls and social-engineering attacks. Second, covert surveillance: a model that tracks multiple speakers over hours could be repurposed for unauthorized monitoring. Third, addressee manipulation: targeted-response capability could amplify persuasive or coercive content in group conversations.

We adopt three mitigations. First, \texttt{MultiTalkPT/FT} contains no real-speaker identity. Second, the released model checkpoint is gated behind a usage policy that prohibits impersonation, deceptive media, and surveillance applications, mirroring the policies of comparable speech-foundation releases. Third, we publish the engine prompts and filter rules in full, so that downstream users can audit and adjust the conversational distribution they generate.

\section{Licenses}
\label{app:licenses}

All third-party assets were verified against their authoritative sources.

\paragraph{Models and tools.}
Moshiko-7B \cite{kyutai2024moshi} (code MIT, weights CC-BY 4.0), IndexTTS2 \cite{zhou2025indextts2} (Apache-2.0), Montreal Forced Aligner \cite{mcauliffe2017mfa} (MIT), PersonaPlex \cite{nvidia2026personaplex} (code MIT, weights NVIDIA Open Model License), MiniCPM-o 4.5 \cite{yao2024minicpm} (code and weights Apache-2.0), and Qwen3-Omni-30B-A3B-Instruct \cite{xu2025qwen3omni} (Apache-2.0). The judges Gemma-4-31B \cite{gemma4deepmind2026} and Qwen3.5-27B \cite{qwen35} are governed by the Gemma Terms of Use and Apache-2.0 respectively. Gemini 2.5 Pro \cite{comanici2025gemini25pushingfrontier}, used for character-seed normalization, is governed by the Gemini API Additional Terms, which restrict using outputs to train competing generative services. Claude Sonnet 4.6 \cite{anthropic2026sonnet46card}, used for dialogue script synthesis, is governed by Anthropic's Usage Policies and Commercial Terms of Service, which prohibit using outputs to develop competing models or services. As with Gemini, this restriction propagates to our derivatives.

\paragraph{Speech corpora.}
CC-BY 4.0: AMI \cite{carletta2006ami}, ICSI \cite{janin2003icsi}. CC-BY-SA 4.0: CHiME-6 \cite{watanabe2020chime6}, AISHELL-4 \cite{fu2021aishell4}, AliMeeting \cite{yu2022alimeeting}, ASCEND \cite{lovenia2022ascend}. CC-BY-NC-ND 4.0: MagicData-RAMC \cite{yang2022magicdata}. DNS-Challenge noise \cite{dubey2023dns} is distributed under MIT and CC-BY 4.0 with mixed component licenses.

\paragraph{Text seed sources.}
\texttt{dair-ai/emotion} \cite{saravia2018carer} is licensed for educational and research use only and derives from public Twitter/X content. \texttt{a-m-team/AM-DeepSeek-R1-0528-Distilled} \cite{tian2025amdistillation} and \texttt{a-m-team/AM-Qwen3-Distilled} \cite{tian2025amdistillation} are research-only per the a-m-team distillation-series notice. These restrictions propagate to our derivatives.

\paragraph{Released artifacts.}
The MultiTalk data engine is released under Apache-2.0. \datapt and \dataft are released under CC-BY-NC 4.0, dictated by the research-only text seed datasets, and the Gemini API terms. MultiTalkBench is released under CC-BY-SA 4.0. The accompanying scoring scripts, persona prompts, segmentation indices, and judge templates are released under Apache-2.0.

\begin{description}
    \item[\datapt] Bilingual dyadic pre-training corpus, $54.4$\,k hours, parallel-stream audio with codec-frame-level word alignments. \\
    \url{https://huggingface.co/datasets/MultiTalk/MultiTalkPT}
    \item[\dataft] Multi-party fine-tuning corpus, $3.2$\,k hours, parallel-stream audio with role and speaker labels. \\
    \url{https://huggingface.co/datasets/MultiTalk/MultiTalkFT}
    \item[\databench] Long-form, multi-party, bilingual full-duplex evaluation benchmark. \\
    \url{https://huggingface.co/datasets/MultiTalk/MultiTalkBench}
    \item[Code] Data engine, evaluation and scoring scripts. \\
    \url{https://anonymous.4open.science/r/MultiTalk/}
\end{description}

\section{Details of MultiTalkBench}
\label{app:multitalkbench}

\paragraph{Overview.}
MultiTalkBench is the first benchmark to jointly evaluate long, multi-party,
bilingual full-duplex spoken dialogue. Each evaluation \emph{sample}
corresponds to a single \texttt{(meeting, target speaker)} pair: the
system-under-test plays the role of one named participant in an otherwise
human-conducted meeting, while the other participants are reproduced from
ground-truth alignments. A sample is graded along three concerns:
\textbf{(a)~long interactions} ($>$10\,min, with explicit probes for
long-range entity tracking and topic coherence),
\textbf{(b)~one-model-many-user multi-party interaction} (addressee
selection, turn-taking, group awareness), and
\textbf{(c)~English--Chinese bilingual ability}. The dataset (audio,
sentence-level alignments, persona prompts, meeting profiles) is released
on the Hugging Face Hub at
\href{https://huggingface.co/datasets/MultiTalk/MultiTalkBench}{\texttt{MultiTalk/MultiTalkBench}}.
This appendix documents the tasks the benchmark imposes on each sample.

\subsection{Four Roles}
\label{app:mtb:roles}

Spoken meetings are conducted by participants playing one of four
behavioural roles:

\begin{description}
\item[Facilitator.] Opens or closes the meeting,
transitions between agenda items at least three times, allocates speaking
turns, and announces or confirms decisions.
\item[Driver.] Initiates new directions and concrete
proposals more often than they respond, and carries the discussion forward
with substantive content.
\item[Collaborator.] Builds on others' ideas,
integrates opposing positions and lowers friction. The share of extending
utterances exceeds the share of independent initiations.
\item[Evaluator.] Surfaces risks, counter-examples
and constraints. Questioning utterances exceed 35\,\% of their turns.
\end{description}

\subsection{Metric}
\label{app:mtb:rubric}

Each completed sample is scored along 14 or 15 ordinal dimensions
$\in\{1,2,3,4,5\}\cup\{\textsc{n/a}\}$, organised into three groups, and either 5
(Facilitator) or 4 (Driver / Collaborator / Evaluator) role-specific
dimensions, for a rubric total of 27 ordinal dimensions across all roles.

\paragraph{General dimensions (G).}
Applied to every sample regardless of role:
\begin{description}\itemsep0pt
\item[\textbf{G1 -- Clarity and Actionability.}] The utterances are clear
and contain enough concrete information to react, decide, or act on.
\item[\textbf{G2 -- Relevance and Accuracy of Response.}] Replies address what was actually
said, with the right qualifiers and no straw-manning or partial answers.
\item[\textbf{G3 -- Agenda Fit and Rhythm Awareness.}] Contributions match the active
topic and the meeting stage (explore / align / decide / wrap up).
\item[\textbf{G4 -- Contribution to Meeting Progress.}] After the speaker
contributes, the discussion is clearer, more focused, or closer to a
decision rather than circular.
\end{description}

\paragraph{Multi-party dimensions (M).}
The MPIQ (Multi-Person~IQ) axes that distinguish multi-party from dyadic
dialogue:
\begin{description}\itemsep0pt
\item[\textbf{M1 -- Multi-Speaker State Tracking.}] Maintains correct
attribution of statements to speakers across long contexts, including
under $\geq 4$ active participants.
\item[\textbf{M2 -- Differentiated Response.}] Gives distinguishably
different answers to participants with different positions, and does not
paper over disagreement with generic acknowledgements.
\item[\textbf{M3 -- Cross-Speaker Information Integration.}] Merges
information scattered across speakers, and identifies cross-turn
contradictions and latent consensus.
\item[\textbf{M4 -- Multi-Party Disagreement Handling.}] When three or
more parties disagree, identifies each party's core concern and proposes
an integrative path rather than handling only the loudest two-way
conflict.
\item[\textbf{M5 -- Group Awareness.}] Recognises that every utterance is
public to all participants, balances its content across listeners, and
does not strongly endorse one party's controversial proposal in front of
the others.
\item[\textbf{M6 -- Noise Resistance \& Input Priority.}] Avoids being
overridden by the most recent or most repeated input, and explicitly flags
contradictory instructions instead of silently siding.
\end{description}

\paragraph{Role-specific dimensions (F\,/\,D\,/\,C\,/\,E).}
A further 4--5 dimensions are evaluated conditional on the primary role:

\begin{center}
\begin{tabular}{@{}lp{0.74\linewidth}@{}}
\toprule
\textbf{Facilitator}  & F1~Agenda~Momentum,\;
                        F2~Participation~Activation,\;
                        F3~Decision~Convergence,\;
                        F4~Summary~\&~Closure,\;
                        F5~Neutrality~\&~Fairness. \\
\textbf{Driver}       & D1~New-Direction~Initiation,\;
                        D2~Proposal~Actionability,\;
                        D3~Argumentation~\&~Push,\;
                        D4~Pace~Control. \\
\textbf{Collaborator} & C1~Extension~\&~Building,\;
                        C2~Integration,\;
                        C3~Friction~Reduction,\;
                        C4~Supportive~Progress. \\
\textbf{Evaluator}    & E1~Accuracy~of~Challenges,\;
                        E2~Constructive~Criticism,\;
                        E3~Sufficiency~of~Evidence,\;
                        E4~Risk-to-Action~Conversion. \\
\bottomrule
\end{tabular}
\end{center}

\medskip
\noindent
The full per-dimension definitions, evaluation cues, and high/low-score
signals (the strings actually inserted into the judge's system prompt) are
released alongside the code in \texttt{prompt\_builder.py}. We summarise
them here for space.

\subsection{Judging Pipeline}
\label{app:mtb:pipeline}

Each sample is scored by exactly six calls to a judge LLM
(Gemma-4-31B-it). The calls
are organised so that the only sequential dependency is role
identification:
\[
\underbrace{\text{Call 1}}_{\text{role}}
\;\longrightarrow\;
\underbrace{\{\text{2a},\text{2b},\text{2c}\}}_{\text{MPIQ group, parallel}}
\;\cup\;
\underbrace{\{\text{3a},\text{3b}\}}_{\text{dim group, parallel}}.
\]
The transcript that every call sees is built by interleaving the model's
own offline output (timestamped tokens, segmented at silence gaps of
$1.2$\,s) with the ground-truth peer alignments, sorted by start time, and
re-aliased so speakers appear as \texttt{A:}, \texttt{B:}, \dots\ to
discourage memorisation of the persona prompt. Each judge prompt also
carries an explicit reminder that only lines beginning with the target
alias may be used as evidence. The roles, calls, and dimensions covered
are summarised in Table~\ref{tab:judge-calls}.

\begin{table}[t]
\centering
\caption{Six-call judging pipeline per evaluation sample.}
\label{tab:judge-calls}
\begin{tabular}{@{}lll@{}}
\toprule
Call & Group & Dimensions / output \\
\midrule
1   & --       & primary role, secondary role, confidence, turn count \\
2a  & MPIQ     & M1, M2 \\
2b  & MPIQ     & M3, M4 \\
2c  & MPIQ     & M5, M6 \\
3a  & General  & G1, G2, G3, G4 \\
3b  & Role     & 4--5 role-specific dims (F\textsuperscript{*}, D\textsuperscript{*}, C\textsuperscript{*}, or E\textsuperscript{*}) \\
\bottomrule
\end{tabular}
\end{table}

\subsection{Mechanical Participation Score}
\label{app:mtb:coef}

The judge measures the \emph{quality} of the system's contributions. We
need an additional, untrainable signal for whether the system spoke at all
and at the right rate. From the offline alignment we extract a
\emph{participation ratio}
\[
r_{\textsc{sut}} \;=\;
\frac{\text{(last target end)} - \text{(first target start)}}
     {\text{(last meeting end)} - \text{(first meeting start)}}
\;\in\;[0,1],
\]
and the analogous ratio $r_{\textsc{gt}}$ for the human reference speaker
on the same sample. The participation coefficient is
\[
\mathrm{coef} \;=\; \max\!\bigl(0,\; 1 - \lvert r_{\textsc{sut}} - r_{\textsc{gt}} \rvert\bigr).
\]
A model whose presence matches the human reference incurs no penalty
($\mathrm{coef}=1$). A silent model loses approximately $1-r_{\textsc{gt}}$.
A model that monologues against a participatory reference is penalised
symmetrically. This is reported separately as
\textbf{P1}~$=100\cdot\mathrm{coef}\in[0,100]$.

\subsection{Final Score}
\label{app:mtb:score}

Each ordinal dimension score $s\in\{1,\ldots,5,\textsc{n/a}\}$ is normalised to $[0,100]$ by $\tilde{s}=(s-1)/4\cdot 100$, with \textsc{n/a} pinned to the rubric minimum (treated as $\tilde{s}=0$).

Let $\mathcal{S}$ denote the full set of $N=|\mathcal{S}|$ evaluation samples, and let $\mathcal{S}_\rho\subset\mathcal{S}$ denote the subset assigned role $\rho\in\{F,D,C,E\}$, so that $\{\mathcal{S}_\rho\}_\rho$ partitions $\mathcal{S}$. For a single sample, the General and \textsc{mpiq} groups average across all of their sub-dimensions, while the Role-conditional group averages only across the sub-dimensions of that sample's assigned role $\rho(s)$:
\[
G(s)=\tfrac{1}{4}\sum_{i=1}^{4}\tilde{s}_{G_i}(s),\qquad
M(s)=\tfrac{1}{6}\sum_{i=1}^{6}\tilde{s}_{M_i}(s),\qquad
R(s)=\tfrac{1}{|D_{\rho(s)}|}\sum_{i\in D_{\rho(s)}}\tilde{s}_{R_i}(s),
\]
where $D_\rho$ is the set of sub-dimensions defined for role $\rho$ ($|D_F|=5$, $|D_D|=|D_C|=|D_E|=4$).

The General and \textsc{mpiq} group scores reported in Table~2 are sample averages,
\[
\bar G=\tfrac{1}{N}\sum_{s\in\mathcal{S}}G(s),\qquad
\bar M=\tfrac{1}{N}\sum_{s\in\mathcal{S}}M(s),
\]
whereas each per-role column is the role's contribution to the overall sample average,
\[
\bar R_\rho=\tfrac{1}{N}\sum_{s\in\mathcal{S}_\rho}R(s)=\tfrac{|\mathcal{S}_\rho|}{N}\cdot\underbrace{\tfrac{1}{|\mathcal{S}_\rho|}\sum_{s\in\mathcal{S}_\rho}R(s)}_{\text{mean within role }\rho},
\]
so that the Role-conditional group score is exactly the sum of the four per-role columns and equals the sample average of $R(s)$:
\[
\bar R=\sum_{\rho\in\{F,D,C,E\}}\bar R_\rho=\tfrac{1}{N}\sum_{s\in\mathcal{S}}R(s).
\]
The final score combines the three groups equally:
\[
\textsc{FinalScore}=\tfrac{1}{3}\!\left(\bar G+\bar M+\bar R\right)\in[0,100].
\]

Grades are assigned via fixed thresholds:
A+~$\geq 90$, A~$\geq 85$, A$-$~$\geq 80$, B+~$\geq 75$, B~$\geq 70$,
B$-$~$\geq 65$, C+~$\geq 60$, C~$\geq 55$, C$-$~$\geq 50$, F~otherwise.

\newpage
\section*{NeurIPS Paper Checklist}

\begin{enumerate}

\item {\bf Claims}
    \item[] Question: Do the main claims made in the abstract and introduction accurately reflect the paper's contributions and scope?
    \item[] Answer: \answerYes{}
    \item[] Justification: The three contributions claimed in the abstract and Section~\ref{sec:intro} (an open data engine and 57.6k-hour corpus, MultiTalkBench, and a bilingual Moshi-style model) are realized by Section~\ref{sec:methods} and substantiated experimentally in Section~\ref{sec:experiments}.

\item {\bf Limitations}
    \item[] Question: Does the paper discuss the limitations of the work performed by the authors?
    \item[] Answer: \answerYes{}
    \item[] Justification: We discuss limitations in the dedicated paragraph ``Limitations and headroom'' in Section~\ref{sec:main_results} and in the expanded Limitations section (Appendix~\ref{app:limitations}).

\item {\bf Theory assumptions and proofs}
    \item[] Question: For each theoretical result, does the paper provide the full set of assumptions and a complete (and correct) proof?
    \item[] Answer: \answerNA{}
    \item[] Justification: The paper presents an empirical data-engine, benchmark, and model contribution. It contains no formal theorems or proofs.

    \item {\bf Experimental result reproducibility}
    \item[] Question: Does the paper fully disclose all the information needed to reproduce the main experimental results of the paper to the extent that it affects the main claims and/or conclusions of the paper (regardless of whether the code and data are provided or not)?
    \item[] Answer: \answerYes{}
    \item[] Justification: The data engine pipeline is described in Section~\ref{sec:data_engine}. Benchmark construction is described in Section~\ref{sec:multitalkbench}. The two-phase training recipe, with all loss weights, learning rates, schedules, batch size, and step counts, is described in Section~\ref{sec:experiments}. The released corpus and benchmark, together with the code, are sufficient for replication.

\item {\bf Open access to data and code}
    \item[] Question: Does the paper provide open access to the data and code, with sufficient instructions to faithfully reproduce the main experimental results, as described in supplemental material?
    \item[] Answer: \answerYes{}
    \item[] Justification: All assets are submitted as anonymized URLs for review and will be hosted publicly upon de-anonymization. Reproduction commands are documented in the README.

\item {\bf Experimental setting/details}
    \item[] Question: Does the paper specify all the training and test details (e.g., data splits, hyperparameters, how they were chosen, type of optimizer) necessary to understand the results?
    \item[] Answer: \answerYes{}
    \item[] Justification: Section~\ref{sec:experiments} specifies the training and test details.

\item {\bf Experiment statistical significance}
    \item[] Question: Does the paper report error bars suitably and correctly defined or other appropriate information about the statistical significance of the experiments?
    \item[] Answer: \answerNo{}
    \item[] Justification: To address reliability, we instead report Spearman's $\rho$ and Kendall's $\tau$ between human and LLM-judge scores, plus inter-judge agreement (Section~\ref{sec:human_evaluation}, Table~\ref{tab:human-eval}).

\item {\bf Experiments compute resources}
    \item[] Question: For each experiment, does the paper provide sufficient information on the computer resources (type of compute workers, memory, time of execution) needed to reproduce the experiments?
    \item[] Answer: \answerYes{}
    \item[] Justification: Compute for both training phases, data-engine, and benchmark inference are reported in Appendix~\ref{app:compute}.
    
\item {\bf Code of ethics}
    \item[] Question: Does the research conducted in the paper conform, in every respect, with the NeurIPS Code of Ethics \url{https://neurips.cc/public/EthicsGuidelines}?
    \item[] Answer: \answerYes{}
    \item[] Justification: The research conforms to the NeurIPS Code of Ethics.

\item {\bf Broader impacts}
    \item[] Question: Does the paper discuss both potential positive societal impacts and negative societal impacts of the work performed?
    \item[] Answer: \answerYes{}
    \item[] Justification: Positive and negative societal impacts are discussed in Appendix~\ref{app:broader_impacts}.
    
\item {\bf Safeguards}
    \item[] Question: Does the paper describe safeguards that have been put in place for responsible release of data or models that have a high risk for misuse (e.g., pre-trained language models, image generators, or scraped datasets)?
    \item[] Answer: \answerYes{}
    \item[] Justification: Appendix~\ref{app:broader_impacts} details our release safeguards.

\item {\bf Licenses for existing assets}
    \item[] Question: Are the creators or original owners of assets (e.g., code, data, models), used in the paper, properly credited and are the license and terms of use explicitly mentioned and properly respected?
    \item[] Answer: \answerYes{}
    \item[] Justification: All upstream assets are cited in the main text and licensed appropriately in Appendix~\ref{app:licenses}.

\item {\bf New assets}
    \item[] Question: Are new assets introduced in the paper well documented and is the documentation provided alongside the assets?
    \item[] Answer: \answerYes{}
    \item[] Justification: The new assets introduced in this paper are well-documented, with anonymized URLs provided for review.

\item {\bf Crowdsourcing and research with human subjects}
    \item[] Question: For crowdsourcing experiments and research with human subjects, does the paper include the full text of instructions given to participants and screenshots, if applicable, as well as details about compensation (if any)? 
    \item[] Answer: \answerNA{}
    \item[] Justification: The paper does not involve crowdsourcing or research with human subjects.

\item {\bf Institutional review board (IRB) approvals or equivalent for research with human subjects}
    \item[] Question: Does the paper describe potential risks incurred by study participants, whether such risks were disclosed to the subjects, and whether Institutional Review Board (IRB) approvals (or an equivalent approval/review based on the requirements of your country or institution) were obtained?
    \item[] Answer: \answerNA{}
    \item[] Justification: The paper does not involve crowdsourcing or research with human subjects.

\item {\bf Declaration of LLM usage}
    \item[] Question: Does the paper describe the usage of LLMs if it is an important, original, or non-standard component of the core methods in this research? Note that if the LLM is used only for writing, editing, or formatting purposes and does \emph{not} impact the core methodology, scientific rigor, or originality of the research, declaration is not required.
    \item[] Answer: \answerYes{}
    \item[] Justification: LLMs are a core component of our data engine.

\end{enumerate}

\end{document}